\documentclass{article} % For LaTeX2e
\usepackage{iclr2027_conference,times}

\usepackage{amsmath,amsfonts,bm}

\def\eqref#1{equation~\ref{#1}}
\def\1{\bm{1}}

\DeclareMathAlphabet{\mathsfit}{\encodingdefault}{\sfdefault}{m}{sl}
\SetMathAlphabet{\mathsfit}{bold}{\encodingdefault}{\sfdefault}{bx}{n}

\usepackage[utf8]{inputenc} % allow utf-8 input
\usepackage[T1]{fontenc}    % use 8-bit T1 fonts
\usepackage{hyperref}       % hyperlinks
\usepackage{url}            % simple URL typesetting
\usepackage{booktabs}       % professional-quality tables
\usepackage{amsmath}  
\usepackage{amsfonts}       % blackboard math symbols
\usepackage{nicefrac}       % compact symbols for 1/2, etc.
\usepackage{microtype}      % microtypography
\usepackage{xcolor}         % colors
\usepackage{graphicx}
\usepackage{tcolorbox}
\tcbuselibrary{breakable}
\usepackage{soul}
\usepackage{fvextra}

\usepackage{multirow}
\usepackage{colortbl}
\usepackage{array}
\usepackage{wrapfig}
\usepackage{titletoc}

\definecolor{headerblue}{HTML}{C5E0F7}
\definecolor{headerteal}{HTML}{B8E6D0}
\definecolor{rowgray}{HTML}{EEEEEE}
\definecolor{gaingreen}{HTML}{D5F5D5}
\definecolor{lossred}{HTML}{F5D5D5}
\definecolor{linkpink}{HTML}{DC5078}

\newcommand{\nochange}[1]{{\small --}}

\newcommand{\resRandomNubaseAcc}{0.34}

\newcommand{\resLlmSonnetSignorAcc}{0.63}

\newcommand{\resFireNasaAcc}{0.70}

\newcommand{\resOpenScholarSignorAcc}{0.36}
\newcommand{\resOpenScholarSignorAbst}{0.58}

\newcommand{\resOpenScholarNubaseAbst}{0.98}

\newcommand{\resClaudeCodeExtriAcc}{0.82}

\newcommand{\resClaudeCodeNubaseAcc}{0.22}
\newcommand{\resClaudeCodeNubaseAbst}{0.65}

\newcommand{\resCodexSignorAcc}{0.86}

\newcommand{\resCodexNubaseAcc}{0.19}
\newcommand{\resCodexNubaseAbst}{0.59}

\newcommand{\resProClaimSignorAcc}{0.80}

\newcommand{\resProClaimCdbAcc}{0.78}

\newcommand{\resProClaimExtriAcc}{0.72}

\newcommand{\resProClaimNubaseAcc}{0.52}

\newcommand{\resProClaimNasaAcc}{0.66}

\newcommand{\resLadderAbstractsExtriAcc}{0.75}

\newcommand{\resLadderAbstractsNubaseAcc}{0.28}
\newcommand{\resLadderAbstractsNubaseAbst}{0.45}

\newcommand{\resLadderFullTextExtriAcc}{0.83}

\newcommand{\resLadderFullTextNubaseAcc}{0.38}

\newcommand{\resLadderFactsNubaseAbst}{0.15}

\newcommand{\resLadderReactSubclaimsCdbAcc}{0.75}

\newcommand{\resLadderReactSubclaimsNubaseAcc}{0.20}
\newcommand{\resLadderReactSubclaimsNubaseAbst}{0.70}
\newcommand{\resLadderReactSubclaimsNasaAcc}{0.43}

\newcommand{\resJudgeFortyConstantExtriAcc}{0.88}

\newcommand{\resJudgeFortyConstantNasaAcc}{0.50}

\newcommand{\resJudgeFortyRandomExtriAcc}{0.88}

\newcommand{\resJudgeFortyRandomNasaAcc}{0.48}

\newcommand{\resJudgeFortyQwenSignorAcc}{0.63}

\newcommand{\resJudgeFortyQwenSignorIter}{2.73}

\newcommand{\resJudgeFortyHaikuSignorIter}{2.20}

\newcommand{\resJudgeFortyHaikuNubaseAcc}{0.58}

\newcommand{\resJudgeFortySonnetSignorIter}{2.20}

\newcommand{\resJudgeFortySonnetExtriAcc}{0.90}

\newcommand{\resJudgeFortyTrainedSignorAbst}{0.05}
\newcommand{\resJudgeFortyTrainedSignorIter}{3.20}

\newcommand{\resJudgeFortyTrainedExtriAcc}{0.85}

\newcommand{\resJudgeFortyTrainedNubaseAcc}{0.50}

\newcommand{\resJudgeFortyTrainedNasaAbst}{0.05}

\newcommand{\resAblJudgeTrainedAcc}{0.80}

\newcommand{\resAblJudgeTrainedMacroFone}{0.85}
\newcommand{\resAblJudgeTrainedIter}{3.21}

\newcommand{\resAblJudgeTrainedRefR}{0.78}

\newcommand{\resAblJudgeQwenAcc}{0.71}

\newcommand{\resAblJudgeQwenMacroFone}{0.78}
\newcommand{\resAblJudgeQwenIter}{2.63}

\newcommand{\resAblJudgeQwenRefR}{0.65}

\newcommand{\resAblJudgeHaikuAcc}{0.76}

\newcommand{\resAblJudgeHaikuMacroFone}{0.81}
\newcommand{\resAblJudgeHaikuIter}{2.48}

\newcommand{\resAblJudgeHaikuSupR}{0.92}

\newcommand{\resAblJudgeHaikuRefR}{0.67}

\newcommand{\resMcnStmtOneP}{$p = 0.022$}
\newcommand{\resMcnJudgeHaikuA}{0.83}
\newcommand{\resMcnJudgeHaikuB}{0.74}

\newcommand{\resMcnJudgeSonnetA}{0.83}
\newcommand{\resMcnJudgeSonnetB}{0.80}
\newcommand{\resMcnJudgeSonnetP}{$p = 0.50$}
\newcommand{\resMcnStmtTwoP}{$p = 0.0021$}
\newcommand{\resMcnJudgeFixedA}{0.77}
\newcommand{\resMcnJudgeFixedB}{0.65}

\newcommand{\resMcnStmtThreeP}{$p = 0.024$}
\newcommand{\resMcnFactsAbstractsA}{0.77}
\newcommand{\resMcnFactsAbstractsB}{0.65}

\newcommand{\resMcnFactsFullTextB}{0.70}

\newcommand{\resMcnStmtFourP}{$p = 0.0012$}
\newcommand{\resMcnStateReactA}{0.77}
\newcommand{\resMcnStateReactB}{0.60}

\newcommand{\resMcnStateReactFactsB}{0.63}

\newcommand{\resMcnStmtFiveP}{$p = 0.022$}
\newcommand{\resMcnHarnessClaudeCodeA}{0.73}
\newcommand{\resMcnHarnessClaudeCodeB}{0.68}

\newcommand{\resMcnHarnessCodexB}{0.68}

\definecolor{gapxl}{RGB}{144,210,144}   % >=0.30
\definecolor{gapl}{RGB}{180,225,180}    % 0.20--0.29
\definecolor{gapm}{RGB}{210,238,210}    % 0.10--0.19
\definecolor{gaps}{RGB}{232,246,232}    % 0.00--0.09
\definecolor{gapneg}{RGB}{245,213,213}  % ProClaim behind
\newcommand{\gap}[2]{{\setlength{\fboxsep}{0.8pt}\colorbox{#1}{\tiny #2}}}

\title{Does Recursive Knowledge Construction Help Agents Verify Consensus Claims in Science?}

\author{
  Lun Ai\textsuperscript{1,}\thanks{Equal contribution $^\#$Work done during an internship}\quad
  Yu-Heng Wu\textsuperscript{1,$*$,$^\#$}\quad
  Jonathan Schaul\textsuperscript{2} \quad
  Leonie K\"uchenhoff\textsuperscript{2} \quad
  Denes Turei\textsuperscript{2} \\
  \textbf{Francesco Carli\textsuperscript{1}} \quad
  \textbf{Fabio Petroni\textsuperscript{3}} \quad
  \textbf{Aurelien Dugourd\textsuperscript{1,}}\thanks{Joint supervision}\quad
  \textbf{Julio Saez-Rodriguez\textsuperscript{1,2,$\dagger$}} \\[0.5em]
  \parbox[t]{\textwidth}{\normalfont
    \textsuperscript{1}European Bioinformatics Institute (EMBL-EBI), European Molecular Biology Laboratory, Hinxton, UK.\quad
    \textsuperscript{2}Institute for Computational Biomedicine, Heidelberg University, Faculty of Medicine, and Heidelberg University Hospital, Heidelberg, Germany.\quad
    \textsuperscript{3}EMBL Rome, European Molecular Biology Laboratory, Monterotondo, Italy.\\[0.3em]
    \texttt{\{ail,dugourd,saez\}@ebi.ac.uk}
  } %\quad
}

\iclrfinalcopy % Uncomment for camera-ready version, but NOT for submission.
\begin{document}

\raggedbottom
\maketitle
\lhead{Preprint}

\begin{abstract}
Recent scientific agents build knowledge recursively by retrieving evidence, constructing an intermediate state, and judging whether that state contains sufficient information. These agents are evaluated end-to-end, usually without ground truth, so the effect of each design choice is unknown. Here, we examined two design axes: (1) what the recursion builds (the state of evidence) and (2) what controls it (the judge that decides when to stop). To measure them, we defined consensus scientific claim verification and measured these effects on this task, where a system must establish a claim's consensus stance from the open literature. We built a new consensus claim dataset, ProClaim-eval, including 718 claims with SUPPORT and REFUTE gold labels from five expert-curated databases spanning molecular biology, nuclear physics, and astronomy. Our instrument, ProClaim, persists an evidence state outside the model's context, so each axis can be studied with the other held fixed. Against verification systems and general-purpose agent harnesses, ProClaim ranks among the top three, with a significant advantage over the harnesses. We found that the evidence state improves performance over in-context accumulation, and it also matters what is stored in it. In addition, a trained judge that reads the evidence state performs better than fixed-round stopping. A judge trained separately and never exposed to ProClaim-eval achieves the highest agreement with the gold labels on three of five subsets, where it matches the off-the-shelf Claude Sonnet~4.6 judge and outperforms Claude Haiku~4.5. Our results support recursively maintaining a structured evidence state outside the model's context as an effective design for verifying claims in scientific agents.
\end{abstract}

\section{Introduction}
\label{sec:intro}

\begin{wrapfigure}{r}{0.5\textwidth}
    \vspace{-10pt}
    \centering
\includegraphics[width=0.45\textwidth]{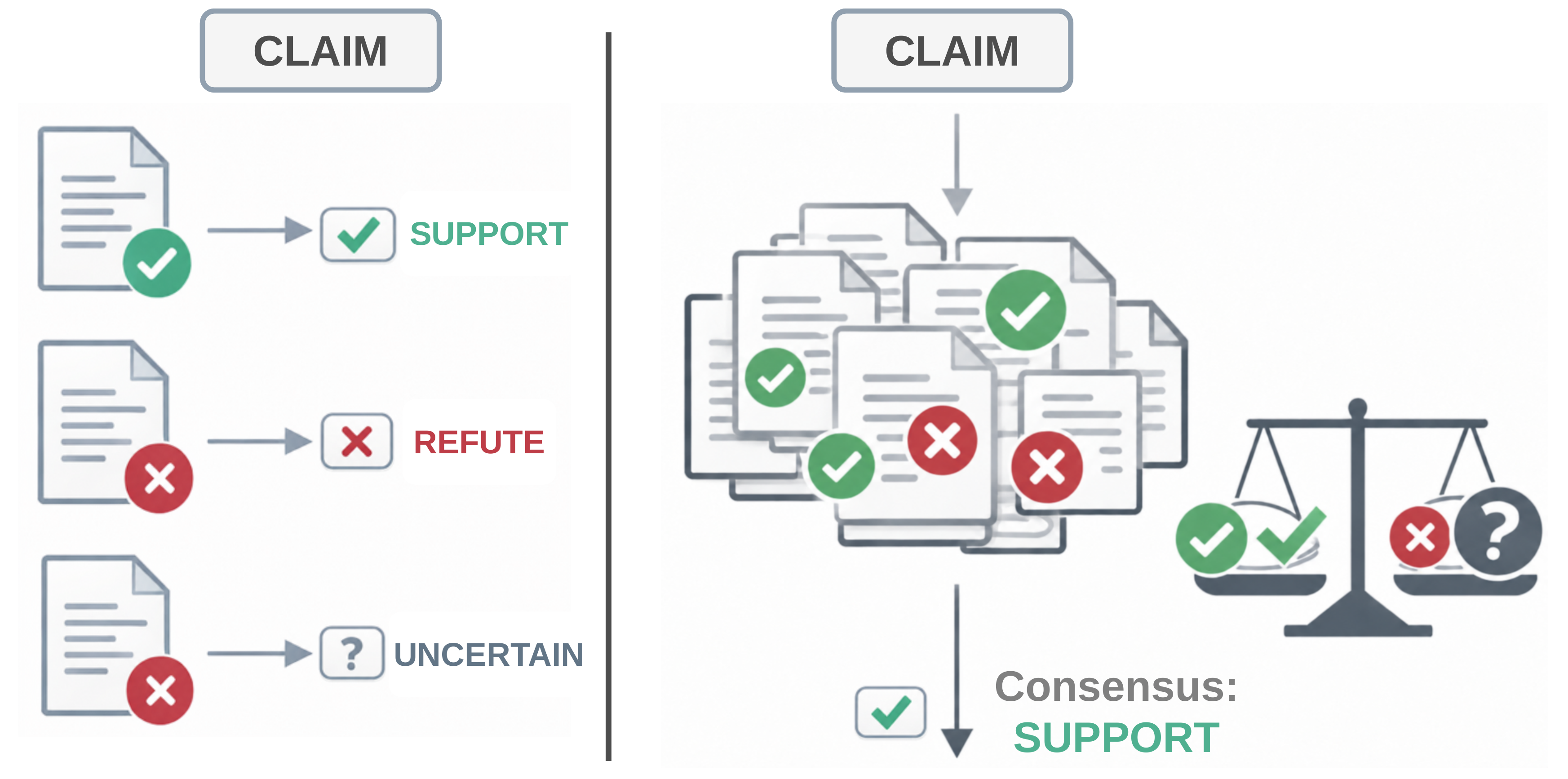}
    \caption{Document-guided vs.\ consensus scientific claim verification.}
    \label{fig:motivation}
    % \vspace{-30pt}
    \vspace{-10pt}
\end{wrapfigure}

Scientific agents built on large language models (LLMs) can now retrieve, synthesise, and reason over the research literature at scale \citep{bachman_automated_2023,wood_empowering_2026}. Many agentic systems \citep{ghareeb_robin_2025,skarlinski_language_2024,mitchener_kosmos_2025,gottweis_accelerating_2026} execute a common loop, which constructs a record of evidence and judges whether the record suffices for a given task, and if not, uses the record to direct the next retrievals. The loop is recursive because each step's record is the next step's input \citep{zhang_recursive_2025}. However, these agents contain many components and are evaluated only on final outputs, so effectiveness cannot be attributed to any single part. Whether \emph{what the recursion builds} and \emph{what controls it} actually help is therefore assumed rather than carefully examined. We call these the two design axes of \emph{recursive knowledge construction}.

To measure each axis, we require a task that exercises the loop and has established consensus. Scientific claim verification is the natural candidate. Claims are the basic unit of literature synthesis and hypothesis generation, and building the consensus on a claim is essential to counter research misinformation \citep{deng_next_2025,diggelmann_climate-fever_2020,van_der_linden_misinformation_2022,west_misinformation_2021} and to uncover gaps that may drive new discovery \citep{asai_synthesizing_2026,vladika_scientific_2023}. Existing datasets \citep{diggelmann_climate-fever_2020,reisle_evaluating_2025,saakyan_covid-fact_2021,sarrouti_evidence-based_2021,wadden_fact_2020,wadden_scifact-open_2022}, however, have reduced verification to reading comprehension over a single retrieved passage \citep{vladika_comparing_2024}. This document-guided setting (Figure \ref{fig:motivation}) has inherent constraints, as verdicts grounded in a limited context cannot reflect the actual scientific stance of claims across the broader literature \citep{deng_next_2025,kumar_sciclaimhunt_2025,skarlinski_language_2024}. We therefore define \emph{consensus claim verification} as an open-domain \citep{vladika_comparing_2024}, claim-level task with labels reflecting the epistemic state of the literature. Solving it requires retrieving, organising and reconciling sparse and conflicting evidence across many sources \citep{vladika_step-by-step_2025,wang_factcheck-bench_2024}, which functionally demands a recursive agentic loop.

% Existing claim verification datasets \citep{reisle_evaluating_2025,saakyan_covid-fact_2021,sarrouti_evidence-based_2021,wadden_fact_2020,wadden_scifact-open_2022} offer only passage-level verdict annotations, as curating consensus stances demands substantial expertise and effort \citep{guo_survey_2022,wadden_fact_2020,wang_factcheck-bench_2024}. 

\begin{figure}[t]
\centering
\includegraphics[width=\linewidth]{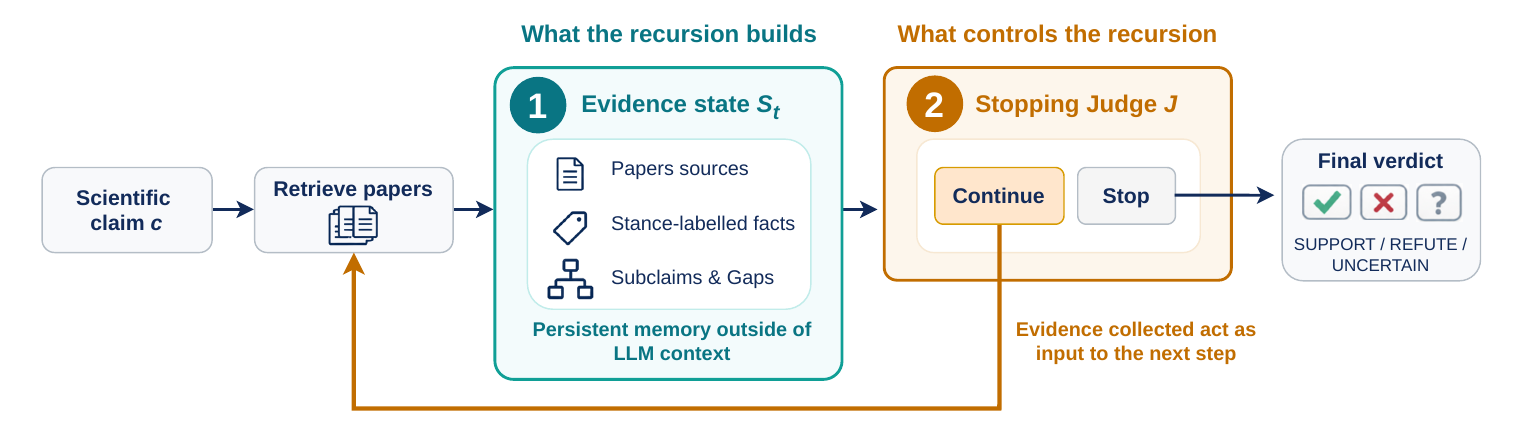}
\caption{Recursive knowledge construction for claim verification. We examine two design choices in Section \ref{sec:experiments}: (1) an evidence state that the next retrieval is conditioned on, (2) a judge that takes the state as input and decides either to continue or terminate the construction, after which the verdict is given from the final state. }
\label{fig:rkc}
\end{figure}

To make this setting evaluable, we release \textbf{ProClaim-eval}, a new dataset of 718 consensus claims built from five actively curated expert databases spanning molecular biology \citep{LoSurdo2025,liu_connectomedb2025_2026,vazquez_extri_2022}, nuclear structure \citep{tuli_evaluated_1996} and astronomy \citep{akeson_nasa_2013,christiansen_nasa_2025}. Prior claim verification datasets offer only passage-level annotations because curating consensus stances demands substantial expertise \citep{guo_survey_2022,wadden_fact_2020,wang_factcheck-bench_2024}. The chosen databases are actively maintained, so their entries track the current domain consensus rather than the stance of any one paper. Gold labels are \textsc{support} or \textsc{refute}, and a system may additionally answer \textsc{uncertain}, which we treat as an abstention. To vary the two design choices in recursive knowledge construction (Figure \ref{fig:rkc}), we built \textbf{ProClaim}, a minimal open-source instrument that maintains an evolving evidence state and a replaceable judge. Findings from our instrument only generalise if it performs on par with systems in use, so we compared ProClaim with state-of-the-art verification systems \citep{xie_fire_2025,asai_synthesizing_2026,skarlinski_language_2024} and general-purpose agent harnesses (CODEX and Claude Code).

% As a result, claim-level verification datasets remain limited and underdeveloped, even within the biomedical domain where they are most prevalent \citep{wright_generating_2022}.

Our work makes three main contributions:
\begin{itemize}
    \item \textbf{We formulate consensus scientific claim verification and release ProClaim-eval} as a test domain for recursive knowledge construction, distinct from document-guided verification. ProClaim-eval contains 718 claims from five expert-curated databases across three domains, with binary gold labels. On ProClaim-eval, ProClaim ranks among the top three systems on four subsets, with a significant advantage over the Claude Code and Codex agent harnesses.
    \item \textbf{Recursively built evidence state beats in-context accumulation, and what it stores matters.} On ProClaim-eval, ProClaim outperforms a ReAct agent that accumulates the retrieved papers only in its context, with the same LLM backbone and retrieval tools. Using ProClaim-retrieved papers, we show that the extracted facts in the state give better performance than abstracts or full text as input.
    \item \textbf{A lightweight trained judge can transfer and replace off-the-shelf LLM judges.} With ProClaim as the instrument, we found that a trained judge that reads the state is better and uses fewer steps than a fixed round budget. We trained a lightweight MLP judge, fitted only on SciFact-Open and evaluated in transfer, as an alternative to off-the-shelf LLM judges, which matches the Claude Sonnet~4.6 judge and outperforms the Claude Haiku~4.5 judge on three subsets.
\end{itemize}

\section{Document-Guided Versus ``Consensus'' Claim Verification}
\label{sec:document_guided_vs_in_the_wild}

\begin{table}[t]
\centering
\caption{\textbf{Claim diversity in the ProClaim-eval benchmark.}
We show one example claim per subset, with the subset's claim count and \textsc{support} / \textsc{refute} split. The five subsets instantiate five distinct single-sentence templates over three scientific domains with a total of 718 claims.}
\label{tab:claim-diversity}
\includegraphics[width=\linewidth]{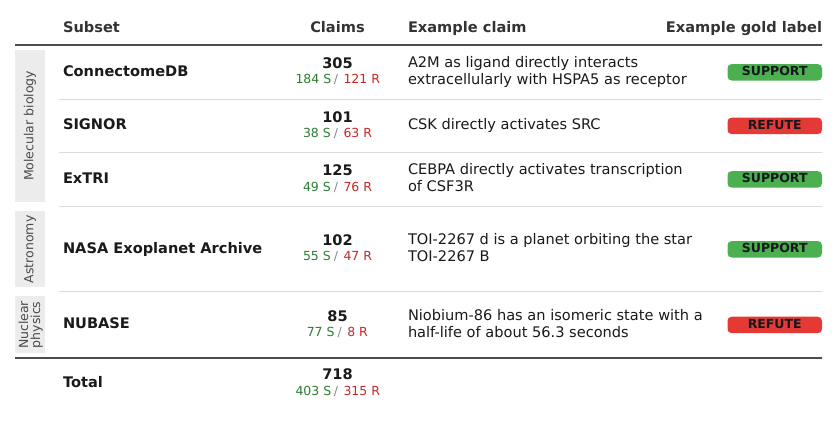}
\vspace{-3em}
\end{table}

Establishing the consensus stance of a scientific claim is essential for uncovering research gaps \citep{asai_synthesizing_2026,vladika_scientific_2023} and curating reliable scientific knowledge \citep{deng_next_2025,diggelmann_climate-fever_2020}. When verifying consensus claims, the verdict emerges from reconciling heterogeneous and conflicting evidence across multiple retrieval passes. Therefore, no single passage can be assumed authoritative for this purpose. Existing benchmarks \citep{wadden_fact_2020,wadden_scifact-open_2022,reisle_evaluating_2025} follow this document-guided paradigm, and are inherently limited. Here, we distinguish these two problem settings:
\begin{itemize}
    \item \textbf{Document-guided verification.} A verdict $v=f(\mathcal{C}, d)$ is associated with a claim $\mathcal{C}$ \emph{and a single passage} $d \in \mathcal{D} = \{d_1,d_2,\ldots,d_N\}$, where $\mathcal{D}$ is a fixed set of documents. The goal is to retrieve the passage given the claim and \emph{classify the pair} $(\mathcal{C}, d)$.
    \item \textbf{Consensus claim verification.} A verdict $v=f(\mathcal{C},\mathcal{D_{\mathcal{L}}})$ is associated with a claim $\mathcal{C}$ \emph{against a set of passages} $\mathcal{D_{\mathcal{L}}} \subseteq \mathcal{L}$, where $\mathcal{L}$ is a large document collection representing the open scientific literature. Since the aim is to \emph{predict the consensus verdict}, the system must determine both which documents to include in $\mathcal{D_{\mathcal{L}}}$ and how many passages are needed.
\end{itemize}
\paragraph{ProClaim-eval.} Existing scientific claim verification datasets contain verdict annotations for each claim--passage pair. This limits evaluation when assembling findings across multiple papers to reach scientific consensus. To address this gap, we created ProClaim-eval (Appendix~\ref{app:dataset_creation}), 718 single-sentence claims from five expert-curated databases spanning three domains (Table~\ref{tab:claim-diversity}). None of the databases provides claim-level verdicts, so each claim's \textsc{support} or \textsc{refute} label was derived from the curators' own record of the underlying evidence. In molecular biology, ConnectomeDB \citep{liu_connectomedb2025_2026} contributes 305 ligand-receptor interactions, labelled \textsc{support} when added in its 2025 release with at least two supporting publications and \textsc{refute} when rejected by its curators from the 2020 release. SIGNOR \citep{LoSurdo2025} contributes 101 expert re-curated claims about signed protein-protein interactions, whose \textsc{refute} claims are interactions removed in later releases or confirmed interactions with a negated sign. ExTRI \citep{vazquez_extri_2022} (part of the CollecTRI database \citep{muller-dott_expanding_2023}) contributes 125 transcription factor-target regulations drawn from expert-read sentences, labelled \textsc{support} when at least five papers agree on the direction of regulation and \textsc{refute} when every curated sentence negates it. In astronomy, the NASA Exoplanet Archive \citep{akeson_nasa_2013,christiansen_nasa_2025} contributes 102 claims that an object is a planet orbiting a given star, labelled \textsc{support} for confirmed planets and \textsc{refute} for published planets later refuted or found to be false positives. In nuclear physics, the NUBASE2020 evaluation \citep{kondev_nubase2020_2021} contributes 85 claims that a nuclide has an isomeric state, labelled \textsc{support} for established states and \textsc{refute} for states reported and later shown not to exist. In ExTRI, NUBASE, and NASA, since their most recent update took place before 2025, we further removed claims that an LLM answers correctly without retrieval, so that a verdict requires retrieving the literature rather than recalling it.

\begin{figure*}[t]
\centering
\includegraphics[width=\linewidth]{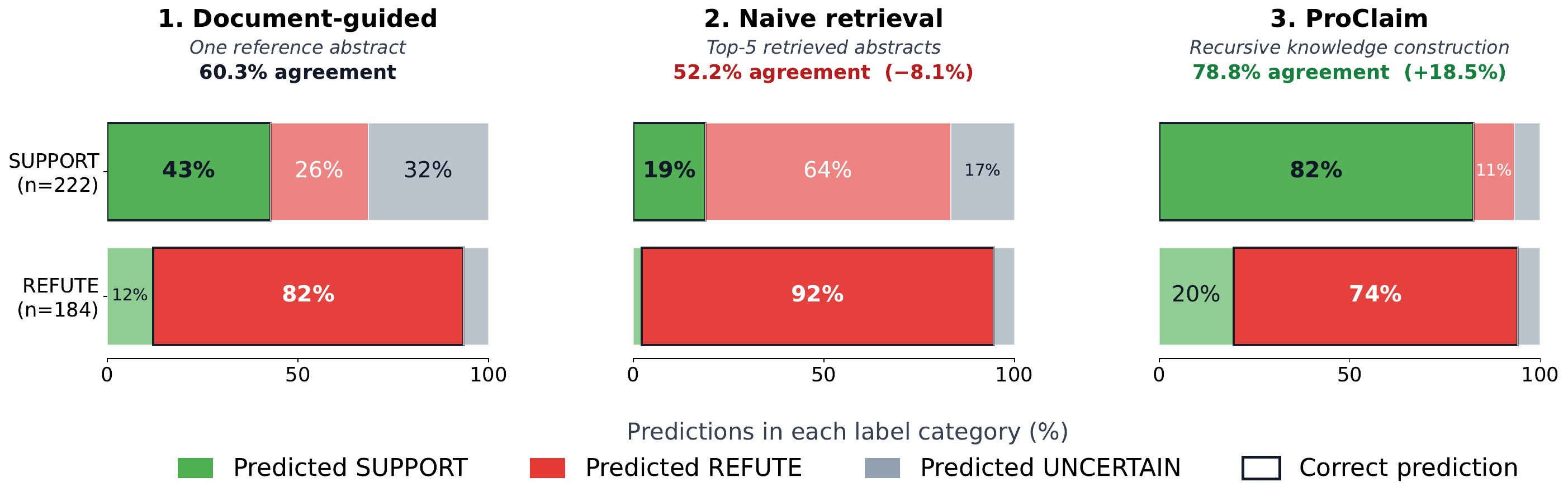}
\caption{
Verdict prediction-label agreement (outlined segments) across three verification settings on the SIGNOR and ConnectomeDB subsets of ProClaim-eval (406 claims), using Claude Sonnet~4.6. Both document-guided and naive retrieval verification do not produce consensus reliably and overproduce \textsc{refute} predictions due to insufficient or noisy information, a limitation that ProClaim addresses. This comparison with ProClaim highlights the benefit of recursive knowledge construction, driven by adaptive retrieval and evidence organisation before deciding the final verdict.
}
\label{fig:sandbox_vs_wild}
\end{figure*}

\paragraph{The need for consensus claim verification.} ProClaim-eval assigns a label to each claim from categories \{\textsc{support}, \textsc{refute}\} (label definitions in Appendix \ref{app:verdict_definitions}). Our analysis in Figure~\ref{fig:sandbox_vs_wild} shows that \textbf{document-guided verification is insufficient for consensus.} With one expert reference abstract as the ``silver'' standard context to guide LLM claim verification, the agreement between verdict predictions and labels is only 60.3\%, which cannot reliably proxy scientific consensus. \textbf{Naive retrieval is insufficient too.} Five retrieved abstracts reduce agreement to 52.2\% and misclassify 24\% \textsc{support} claims compared to using the reference abstract. Importantly, both document-guided verification and static retrieval overproduce \textsc{refute} predictions due to failing to find explicit positive support under limited or noisy evidence, and absence of support after search is treated as REFUTE. 

% In contrast, ProClaim (Section \ref{sec:proclaim}) produces stronger alignment with the \textsc{support} labels and raises agreement to 77.5\%, showing that iterative retrieval, organising, and reconciling improve overall agreement. 

\section{Recursive Knowledge Construction}
\label{sec:rkc}

Our results in Section~\ref{sec:document_guided_vs_in_the_wild} show that verifying scientific consensus cannot be reduced to reading from naively retrieved passages. In this section, we formally define the mechanism \emph{recursive knowledge construction} that recent scientific agents use instead. This led us to build the minimal instrument ProClaim that allows us to vary the two design axes of recursive knowledge construction to examine their effects on consensus verification.

\subsection{Formal Definition}
\label{sec:rkc_def}

We formalise the process shared by recent scientific agents \citep{skarlinski_language_2024,mitchener_kosmos_2025} as a recursion on an explicit evidence state. Let $c$ be a scientific claim and $\mathcal{L}$ a corpus of documents as in the consensus claim verification setting. At step $t$, a verifier maintains an \emph{evidence state} $s_t$ as an intermediate record, which is updated via:
\begin{equation*}
  s_{t+1} = U\big(s_t,\; R(c, s_t)\big), \qquad t = 0, 1, \ldots
\end{equation*}
where $R(c, s_t) \subseteq \mathcal{L}$ decides what to retrieve given what the state already holds, and $U$ updates the state by recording the result. Verification halts at the first round $T$ at which a judgement $J(c, s_T)$ holds, after which the agent produces its output verdict from the final record. Two properties make this process recursive. The evidence state contains the knowledge constructed by the agent, and the state in the previous round becomes the input to the next. The state also guides both where the agent looks next and when it should terminate. 

% I want to say here to establish the connection with Recursive LM work, and recent work in FIRE that uses confidence to stop. 

Two design choices are key in the definition, and they are the objects of study in this paper.
\begin{itemize}
    \item \textbf{Evidence state.} In contrast to claim verification systems such as FIRE \citep{xie_fire_2025}, SAFE \citep{wei_long-form_2024} and OpenScholar \citep{asai_synthesizing_2026}, which keep a record of the growing prompt context, in scientific agents, for example, PaperQA2 \citep{skarlinski_language_2024} and Kosmos \citep{mitchener_kosmos_2025}, the state is explicitly constructed as structure persisted outside the context.
    \item \textbf{Judgement.} The verification process needs a stopping rule. This could be a fixed number of rounds \citep{asai_synthesizing_2026} or a function of the retrieved evidence \citep{xie_fire_2025,asai_self-rag_2024} based on an LLM judgment of whether to keep searching.
\end{itemize}

We implemented a minimal instrument agent, ProClaim, to perform recursive knowledge construction and examine the effects of those two design choices. The experiments therefore varied one axis with the other fixed in ProClaim. 

\subsection{The Evidence State}
\label{sec:evidence_state}

We designed the evidence state as a persistent JSON outside the LLM context. The state records retrieved papers and claim-relevant facts linked to their source paper, with a stance in $\{\textsc{support}, \textsc{refute}, \textsc{neutral}\}$.  We followed the CodeAct \citep{wang_executable_2024} approach, so at each step, the agent writes a short Python program. This program is composed of operations to interact with the state from a small library, including fact extraction with stance labelling from abstracts or full text, subclaim decomposition, and the judge call. 

This state design is minimal and preserves agentic behaviour rather than executing a fixed pipeline. Because the agent needs to access the state through explicit programmatic operations, its content is not compressed or paraphrased as steps accumulate, unlike evidence folded into the growing prompt that dilutes as the context grows. The agent's context is the execution log of the operations it has run and their printed results. Because the query at round $t+1$ is issued after reading the state, retrieval is conditioned on the accumulated evidence as we have defined in Section \ref{sec:rkc_def}. 

% Section~\ref{sec:exp_state} compares this code-composed construction against a fixed pipeline over the same operations, and the typed state against accumulating the same abstracts or full texts in context; implementation details are in Appendix~\ref{app:evidence_programming_impl}.

\subsection{The Judge}
\label{sec:judge}

Section~\ref{sec:document_guided_vs_in_the_wild} shows a characteristic failure in retrieval tasks \citep{xie_fire_2025,jeong_adaptive-rag_2024,asai_self-rag_2024}, where claims require aggregating multiple documents, and an absence of support after a shallow retrieval leads to more \textsc{Refute}. Stopping too early yields a verdict from incomplete or one-sided evidence, but more retrieval on claims can introduce noise or conflict. Following the notion of \emph{sufficient context} \citep{Joren2024SufficientCA}, we therefore wanted the judge to score a property of the evidence state itself. We designed the judge as a scorer $J(c, s_t) \in [0,1]$. The agent stops when $J(c, s_t) \geq \delta$ or reaches a maximum iteration $T_{\max}$ for practical constraints. We implemented $J$ in two ways that consume the same state and the same $\delta = 0.8$, with a cap $T_{\max} = 4$.

\paragraph{Trained judge.} 

We trained a lightweight 2-layer MLP (hidden dimensions 64 and 32) and exposed it to the agent as a tool, so the stopping signal is external rather than a self-assessment. We prepared training data using SciFact-Open \citep{wadden_scifact-open_2022}, which labels stance at the paper level for 279 claims. Papers sharing claims were grouped into the same evidence states and used a minority ratio to annotate sufficiency labels (Appendix \ref{app:sufficiency_dataset}). Per-paper features were extracted from frozen, off-the-shelf encoders, then pooled across all papers in the evidence state into one global representation. We computed the cosine similarity from SBERT \citep{reimers2019sentence} between claim and paper text. This distinguishes states with actually relevant evidence from states with off-topic papers. NLI stance probabilities (entailment, contradiction, neutral) come from a DeBERTa NLI model \citep{he2021debertav3} applied to the claim against the most relevant text chunk from each paper. This acts as a soft, per-paper stance signal. Together with entity coverage and bibliometric metadata, these are pooled into a 10-dimensional global vector that feeds the MLP. We include implementation details in Appendix~\ref{app:sufficiency_classifier_implementation}.

\paragraph{Off-the-shelf judge.} An LLM is prompted zero-shot with the claim and a rendering of $s_t$ (paper titles, abstracts, the per-paper meta-data, and similarity scores from above) and asked to return a sufficiency score, with the definition above stated in the prompt (Appendix~\ref{app:llm_sufficiency_prompt}). 

% We use Claude Haiku~4.5 and, in Section~\ref{sec:exp_judge}, Claude Sonnet~4.6. 

\section{Experiments}
\label{sec:experiments}

\textbf{Shared setting.} Unless otherwise noted, all variations were evaluated across five subsets of the ProClaim-eval dataset. We report \emph{agreement}, the fraction of all claims whose predicted verdict matches the gold label, as in Section~\ref{sec:document_guided_vs_in_the_wild}. A system may answer \textsc{uncertain} to abstain, and we compute the \emph{abstention rate} as the fraction of claims answered \textsc{uncertain}. Every agentic system, including ProClaim, used Claude Sonnet~4.6 with thinking disabled and temperature 0. Paper search was PubMed and Semantic Scholar (S2) on the three biology subsets and S2 alone on non-biology subsets, except for LLM-only baselines where models answer from parameter knowledge. The fact extraction in ProClaim was done by Qwen~3.5 9B. For each finding, we compared two systems on the same claims with a paired exact McNemar test, pooled over subsets, and report Holm-adjusted $p$-values (Appendix~\ref{app:significance}).

% Step budgets per system are listed in Appendix Table~\ref{app:tab:iclr_budget}.

\begin{table}[t]
\centering
\small
\caption{Agreement (Agr.) and abstention rate (Abst.) per ProClaim-eval subset. The shaded figure beside each agreement is ProClaim's agreement minus that row's, on the same subset; green where ProClaim is ahead, red where it is behind. The highest agreement per subset is underlined.}
\vspace{2pt}
\setlength{\tabcolsep}{2.4pt}
\renewcommand{\arraystretch}{1.12}
\begin{tabular}{@{} l | c c | c c | c c | c c | c c @{}}
\toprule
 & \multicolumn{2}{c}{\cellcolor{headerteal}\textbf{SIGNOR}} & \multicolumn{2}{c}{\cellcolor{headerteal}\textbf{ConnectomeDB}} & \multicolumn{2}{c}{\cellcolor{headerteal}\textbf{ExTRI}} & \multicolumn{2}{c}{\cellcolor{headerteal}\textbf{NUBASE}} & \multicolumn{2}{c}{\cellcolor{headerteal}\textbf{NASA}} \\
\textbf{Method} & Agr.$\uparrow$ & Abst. & Agr.$\uparrow$ & Abst. & Agr.$\uparrow$ & Abst. & Agr.$\uparrow$ & Abst. & Agr.$\uparrow$ & Abst. \\
\midrule
\textit{Random} & 0.32 & 0.35 & 0.34 & 0.34 & 0.32 & 0.34 & 0.34 & 0.34 & 0.31 & 0.35 \\
\midrule
\multicolumn{11}{@{}l}{\textit{LLM Only}} \\[2pt]
Claude Sonnet 4.6 & 0.63\,\gap{gapm}{+0.17} & 0.26 & 0.53\,\gap{gapl}{+0.26} & 0.25 & 0.21\,\gap{gapxl}{+0.51} & 0.66 & 0.05\,\gap{gapxl}{+0.47} & 0.95 & 0.07\,\gap{gapxl}{+0.59} & 0.89 \\
Gemini 2.5 Flash & 0.59\,\gap{gapl}{+0.21} & 0.26 & 0.57\,\gap{gapl}{+0.21} & 0.21 & 0.23\,\gap{gapxl}{+0.49} & 0.56 & 0.31\,\gap{gapl}{+0.21} & 0.07 & 0.46\,\gap{gapm}{+0.20} & 0.03 \\
\midrule
\multicolumn{11}{@{}l}{\textit{Static Retrieval, S2 Top 5}} \\[2pt]
Claude Sonnet 4.6 & 0.66\,\gap{gapm}{+0.14} & 0.10 & 0.48\,\gap{gapxl}{+0.30} & 0.12 & 0.60\,\gap{gapm}{+0.12} & 0.06 & 0.13\,\gap{gapxl}{+0.39} & 0.32 & 0.36\,\gap{gapl}{+0.29} & 0.55 \\
Gemini 2.5 Flash & 0.64\,\gap{gapm}{+0.16} & 0.06 & 0.51\,\gap{gapl}{+0.27} & 0.04 & 0.62\,\gap{gapm}{+0.10} & 0.03 & 0.18\,\gap{gapxl}{+0.34} & 0.02 & 0.38\,\gap{gapl}{+0.27} & 0.51 \\
\midrule
\multicolumn{11}{@{}l}{\textit{Adaptive Retrieval}} \\[2pt]
ReAct & 0.78\,\gap{gaps}{+0.02} & 0.10 & 0.70\,\gap{gaps}{+0.08} & 0.08 & 0.74\,\gap{gapneg}{-0.02} & 0.16 & 0.18\,\gap{gapxl}{+0.34} & 0.71 & 0.55\,\gap{gapm}{+0.11} & 0.39 \\
FIRE & 0.75\,\gap{gaps}{+0.05} & 0.06 & 0.50\,\gap{gapl}{+0.29} & 0.14 & 0.73\,\gap{gapneg}{-0.01} & 0.09 & 0.08\,\gap{gapxl}{+0.44} & 0.12 & \underline{0.70}\,\gap{gapneg}{-0.04} & 0.07 \\
OpenScholar & 0.36\,\gap{gapxl}{+0.45} & 0.58 & 0.50\,\gap{gapl}{+0.28} & 0.32 & 0.22\,\gap{gapxl}{+0.50} & 0.73 & 0.02\,\gap{gapxl}{+0.49} & 0.98 & 0.03\,\gap{gapxl}{+0.63} & 0.96 \\
PaperQA2 & 0.71\,\gap{gaps}{+0.09} & 0.11 & 0.70\,\gap{gaps}{+0.08} & 0.08 & 0.79\,\gap{gapneg}{-0.07} & 0.13 & 0.19\,\gap{gapxl}{+0.33} & 0.33 & 0.69\,\gap{gapneg}{-0.03} & 0.17 \\
\midrule
\multicolumn{11}{@{}l}{\textit{General-Purpose Agents}} \\[2pt]
Claude Code & 0.83\,\gap{gapneg}{-0.03} & 0.04 & 0.74\,\gap{gaps}{+0.05} & 0.05 & \underline{0.82}\,\gap{gapneg}{-0.10} & 0.10 & 0.22\,\gap{gapl}{+0.29} & 0.65 & 0.55\,\gap{gapm}{+0.11} & 0.33 \\
Codex & \underline{0.86}\,\gap{gapneg}{-0.06} & 0.04 & 0.74\,\gap{gaps}{+0.05} & 0.03 & 0.79\,\gap{gapneg}{-0.07} & 0.10 & 0.19\,\gap{gapxl}{+0.33} & 0.59 & 0.57\,\gap{gaps}{+0.09} & 0.28 \\
\midrule
\textbf{ProClaim} & 0.80 & 0.12 & \underline{0.78} & 0.05 & 0.72 & 0.14 & \underline{0.52} & 0.36 & 0.66 & 0.14 \\
\bottomrule
\end{tabular}

\label{tab:main_results}
\end{table}

\subsection{How Do Verifiers and Agents Perform on Consensus Claim Verification?}
\label{sec:main_results}

\textbf{Settings.} \emph{Random} draws each verdict uniformly from \textsc{support}, \textsc{refute} and \textsc{uncertain}, averaged over 10 draws, as a chance floor. \emph{LLM-only} makes one call with no retrieved context. \emph{Static retrieval} retrieves the top-5 S2 abstracts for the claim and asks the same LLM for a verdict in one shot. \emph{ReAct} \citep{yao_react_2022} runs a thought-action-observation loop over the paper-search tools. \emph{FIRE} \citep{xie_fire_2025} is a claim verifier built around web search, and we updated its prompts to adopt the paper-search backend as described above. FIRE stops when the LLM reports sufficient confidence. Both ReAct and FIRE were capped at 20 steps, with a budget matching ProClaim's average number of retrievals. \emph{OpenScholar} \citep{asai_synthesizing_2026} retrieves from the same paper-search sources and refines its answer with LLM feedback. We rephrased claims as questions to use OpenScholar’s native LLM feedback pipeline. \emph{PaperQA2} \citep{skarlinski_language_2024} keeps a ranked store of relevance-scored evidence summaries outside the model's context and answers from the highest-ranked ones. It stops when the LLM decides to answer. We replaced its local paper index with the same paper-search tools and open-access full text, and used our verdict defintions, and capped it at 20 steps and 45 minutes per claim. \emph{Claude Code} and \emph{Codex} were general-purpose agent harnesses given the same paper-search tools with web search disabled. To bound run time, Claude Code was capped at 60 turns and Codex at 45 minutes per claim. Every system received the same label definitions (Appendix~\ref{app:verdict_definitions}).

% Most key results for the paper first.
\textbf{Results.} Table~\ref{tab:main_results} reports agreement and abstention per subset. \textbf{First}, ProClaim has the highest agreement on ConnectomeDB (\resProClaimCdbAcc{}) and NUBASE (\resProClaimNubaseAcc{}), trails Codex on SIGNOR (\resProClaimSignorAcc{} vs.\ \resCodexSignorAcc{}), and trails Claude Code on ExTRI (\resProClaimExtriAcc{} vs.\ \resClaudeCodeExtriAcc{}) and FIRE on NASA (\resProClaimNasaAcc{} vs.\ \resFireNasaAcc{}). 
\textbf{Second}, static retrieval scores below LLM-only on ConnectomeDB with both LLMs and on NUBASE and NASA with Gemini~2.5 Flash, showing that naively retrieving the top-5 papers or snippets does not add reliable benefit over parametric knowledge. % CHECK direction
\textbf{Third}, adaptive retrieval improves on static retrieval for the systems that commit to a verdict. ReAct and FIRE exceed static retrieval on almost every subset. OpenScholar abstains on most claims on every subset except ConnectomeDB (from \resOpenScholarSignorAbst{} on SIGNOR to \resOpenScholarNubaseAbst{} on NUBASE) and scores below LLM-only Sonnet~4.6 on SIGNOR (\resOpenScholarSignorAcc{} vs.\ \resLlmSonnetSignorAcc{}). % CHECK direction
\textbf{Fourth}, the SOTA harnesses are strongest on the biology subsets. Codex has the highest SIGNOR agreement (\resCodexSignorAcc{}) and Claude Code the highest ExTRI agreement (\resClaudeCodeExtriAcc{}), but on NUBASE both stay below the random baseline (\resClaudeCodeNubaseAcc{} and \resCodexNubaseAcc{} vs.\ \resRandomNubaseAcc{}), abstaining on \resClaudeCodeNubaseAbst{} and \resCodexNubaseAbst{} of claims. % CHECK direction

\begin{tcolorbox}[
  colback=gray!10,
  colframe=gray!10,
  coltitle=white,
  title={Finding 1: Recursive construction is consistently strong},
  colbacktitle=black!50!red,
  boxrule=0pt,
  left=10pt,
  right=10pt,
  top=4pt,
  bottom=4pt
]
% CHECK consistent with the main result table, think about a title for the box
Recursive knowledge construction shows strong performance on ProClaim-eval. Our instrument ProClaim ranks amongst the top 3 on four subsets, with a significant advantage against harnesses (pooled agreement \resMcnHarnessClaudeCodeA{} vs.\ \resMcnHarnessClaudeCodeB{} for Claude Code and \resMcnHarnessCodexB{} for Codex, \resMcnStmtFiveP{}). 
\end{tcolorbox}

\subsection{What Does the Evidence State Contribute?}
\label{sec:exp_state}

Without an evidence state, a verifier reads what it retrieves as raw text accumulating in its context, and it reaches the verdict from that text. The evidence state (Section~\ref{sec:evidence_state}) instead stores what was read as relevant facts, each linked to its source paper and labelled with a stance towards the claim. Stored this way, the evidence can be read in a distilled form rather than re-read as whole papers, and it persists across rounds rather than diluting as the context grows. We tested whether this representation is a better input to the verdict than the papers it was extracted from, holding the retrieved papers fixed.

\begin{table}[t]
\centering
\small
\caption{Effect of the evidence state. The same LLM (Claude Sonnet~4.6) reads the different information retrieved in the evidence state of ProClaim, against no retrieval. Every row is scored on the same randomly sampled 40 claims per subset. We show the agreement (Agr.) and abstention rate (Abst.) per subset, and the highest agreement per subset is underlined.}
\label{tab:evidence_state}
\vspace{2pt}
\setlength{\tabcolsep}{3pt}
\renewcommand{\arraystretch}{1.12}
\begin{tabular}{@{} l | c c | c c | c c | c c | c c @{}}
\toprule
 & \multicolumn{2}{c}{\cellcolor{headerteal}\textbf{SIGNOR} (40)} & \multicolumn{2}{c}{\cellcolor{headerteal}\textbf{ConnectomeDB} (40)} & \multicolumn{2}{c}{\cellcolor{headerteal}\textbf{ExTRI} (40)} & \multicolumn{2}{c}{\cellcolor{headerteal}\textbf{NUBASE} (40)} & \multicolumn{2}{c}{\cellcolor{headerteal}\textbf{NASA} (40)} \\
\textbf{Evidence \& Methods} & Agr.$\uparrow$ & Abst. & Agr.$\uparrow$ & Abst. & Agr.$\uparrow$ & Abst. & Agr.$\uparrow$ & Abst. & Agr.$\uparrow$ & Abst. \\
\midrule
No retrieval & 0.63 & 0.30 & 0.70 & 0.23 & 0.28 & 0.65 & 0.10 & 0.90 & 0.08 & 0.88 \\
Abstracts & 0.78 & 0.13 & 0.83 & 0.10 & 0.75 & 0.25 & 0.28 & 0.45 & 0.63 & 0.23 \\
Full text & 0.80 & 0.08 & 0.83 & 0.10 & 0.83 & 0.10 & 0.38 & 0.28 & 0.65 & 0.18 \\
Facts & 0.80 & 0.10 & 0.93 & 0.08 & \underline{0.88} & 0.05 & \underline{0.50} & 0.15 & 0.73 & 0.08 \\
Facts + abstracts & 0.80 & 0.08 & 0.93 & 0.05 & 0.83 & 0.13 & \underline{0.50} & 0.15 & \underline{0.75} & 0.03 \\
\midrule
\multicolumn{11}{@{}l}{\textit{No evidence state, ReAct}} \\[2pt]
+ subclaims - state & 0.78 & 0.23 & 0.75 & 0.15 & 0.83 & 0.15 & 0.20 & 0.70 & 0.43 & 0.50 \\
+ subclaims + facts - state & 0.73 & 0.20 & 0.78 & 0.20 & \underline{0.88} & 0.13 & 0.23 & 0.65 & 0.55 & 0.38 \\
\midrule
\textbf{ProClaim} & \underline{0.83} & 0.05 & \underline{0.95} & 0.03 & 0.85 & 0.10 & \underline{0.50} & 0.30 & 0.73 & 0.05 \\
\bottomrule
\end{tabular}

\end{table}

\textbf{Settings.} We varied what the recursion builds while holding the backbone and the retrieval tools fixed, in two ways. We first took the evidence ProClaim retrieved for each claim and gave it once to the same LLM with the same prompt, packaged as retrieved abstracts, full texts where available, the stance-labelled facts of ProClaim's evidence state, or facts plus abstracts. \emph{No retrieval} is the LLM-only baseline. Every other settings gives the final state ProClaim built for the LLM in a single call, so these rows differ only in the content represented. \emph{ProClaim} is the full loop of Table~\ref{tab:main_results}, which builds those states. As a control with no evidence state, \emph{ReAct + subclaims $-$ state} gives ReAct ProClaim's subclaims and the same retrieval tools for 20 steps, and accumulates the papers it reads in its context. \emph{ReAct + subclaims + facts $-$ state} adds ProClaim's fact extraction to this control, where every paper a search returns is passed, as full text where available, to the same extraction model (Qwen 3.5~9B) and prompt, and ReAct sees the extracted facts with each result. Every row was scored on randomly sampled 40 claims per subset.

% The \emph{construction comparison} (Appendix Table~\ref{app:tab:iclr_evidence_state}) sets ProClaim's code-composed construction against ReAct, a fixed tool-calling loop given the same paper-search tools, at 10, 20 and 40 steps.

\textbf{Results.} \textbf{First}, any retrieved evidence moves agreement well above no retrieval on every subset. \textbf{Second}, with the retrieved papers held fixed, the stance-labelled facts are a better input to the verdict than the papers they were extracted from on four of five subsets. Pooled over the five subsets, facts give higher agreement than both abstracts and full text (\resMcnFactsAbstractsA{} vs.\ \resMcnFactsAbstractsB{} and \resMcnFactsFullTextB{}, \resMcnStmtThreeP{}), and lower abstention on five subsets, most on NUBASE (\resLadderFactsNubaseAbst{} vs.\ \resLadderAbstractsNubaseAbst{} with abstracts). \textbf{Third}, more raw text does not substitute for the facts on our dataset. Full text is equivalent to using abstracts on SIGNOR, ConnectomeDB and NASA. It is ahead of abstracts on ExTRI (\resLadderFullTextExtriAcc{} vs.\ \resLadderAbstractsExtriAcc{}) and NUBASE (\resLadderFullTextNubaseAcc{} vs.\ \resLadderAbstractsNubaseAcc{}) but stays below facts on both.
\textbf{Fourth}, ProClaim has higher agreement than both controls without an evidence state on four subsets. Pooled over the five subsets, its agreement is \resMcnStateReactA{} against \resMcnStateReactB{} for ReAct + subclaims $-$ state and \resMcnStateReactFactsB{} for ReAct + subclaims + facts $-$ state (\resMcnStmtFourP{}). ReAct + subclaims $-$ state is below both raw-paper rows on ConnectomeDB (\resLadderReactSubclaimsCdbAcc{}), NUBASE (\resLadderReactSubclaimsNubaseAcc{}, abstaining on \resLadderReactSubclaimsNubaseAbst{} of claims) and NASA (\resLadderReactSubclaimsNasaAcc{}). ProClaim stays ahead of controls on SIGNOR, ConnectomeDB, NUBASE and NASA.

\begin{tcolorbox}[
  colback=gray!10,
  colframe=gray!10,
  coltitle=white,
  title={Finding 2: The evidence state helps, but it matters what it stores},
  colbacktitle=black!50!red,
  boxrule=0pt,
  left=10pt,
  right=10pt,
  top=4pt,
  bottom=4pt
]
The facts in the evidence state are a better input than the abstracts or full text (pooled agreement \resMcnFactsAbstractsA{} vs.\ \resMcnFactsAbstractsB{} and \resMcnFactsFullTextB{}, \resMcnStmtThreeP{}). This is consistent with the prior finding that LLMs use long contexts unevenly \citep{liu_lost_2024}. An agent that keeps papers only in its context does no better than reading raw papers, and ProClaim outperforms it (pooled \resMcnStateReactA{} vs.\ \resMcnStateReactB{} and \resMcnStateReactFactsB{} for the two controls, \resMcnStmtFourP{}).
\end{tcolorbox}

\subsection{What Does a Trained Sufficiency Judge Add over an Off-the-Shelf One?}
\label{sec:exp_judge}

\begin{table}[t]
\centering
\small
\caption{Effect of the sufficiency judge with a fixed evidence state design, on the same 40 randomly sampled claims per subset. Agreement (Agr.), abstention rate (Abst.) and mean judge calls at stop (Iter.) for each stopping rule, the highest agreement per subset is underlined. }
\label{tab:judge}
\setlength{\tabcolsep}{2.4pt}
\resizebox{\linewidth}{!}{\begin{tabular}{@{} l | c c c | c c c | c c c | c c c | c c c @{}}
\toprule
 & \multicolumn{3}{c}{\cellcolor{headerteal}\textbf{SIGNOR} (40)} & \multicolumn{3}{c}{\cellcolor{headerteal}\textbf{ConnectomeDB} (40)} & \multicolumn{3}{c}{\cellcolor{headerteal}\textbf{ExTRI} (40)} & \multicolumn{3}{c}{\cellcolor{headerteal}\textbf{NUBASE} (40)} & \multicolumn{3}{c}{\cellcolor{headerteal}\textbf{NASA} (40)} \\
\textbf{Judge} & Agr.$\uparrow$ & Abst. & Iter. & Agr.$\uparrow$ & Abst. & Iter. & Agr.$\uparrow$ & Abst. & Iter. & Agr.$\uparrow$ & Abst. & Iter. & Agr.$\uparrow$ & Abst. & Iter. \\
\midrule
\multicolumn{16}{@{}l}{\textit{Stopping rules that ignore the evidence state}} \\[2pt]
Fixed rounds & 0.70 & 0.15 & 3.45 & 0.73 & 0.25 & 3.25 & 0.83 & 0.15 & 3.20 & 0.33 & 0.18 & 2.53 & 0.68 & 0.20 & 3.20 \\
Always continue & 0.75 & 0.13 & 2.68 & 0.75 & 0.15 & 3.15 & 0.88 & 0.08 & 3.40 & 0.55 & 0.15 & 2.65 & 0.50 & 0.33 & 2.60 \\
Random score & 0.70 & 0.20 & 3.13 & 0.90 & 0.08 & 3.33 & 0.88 & 0.08 & 3.15 & 0.53 & 0.23 & 2.10 & 0.48 & 0.28 & 2.73 \\
\midrule
\multicolumn{16}{@{}l}{\textit{Off-the-shelf LLM judges}} \\[2pt]
Qwen3.5-9B & 0.63 & 0.23 & 2.73 & 0.90 & 0.08 & 2.68 & 0.80 & 0.13 & 2.73 & 0.50 & 0.33 & 1.93 & 0.58 & 0.23 & 2.05 \\
Claude Haiku 4.5 & 0.70 & 0.15 & 2.20 & 0.88 & 0.10 & 2.48 & 0.83 & 0.15 & 2.63 & \underline{0.58} & 0.20 & 2.23 & 0.65 & 0.13 & 2.18 \\
Claude Sonnet 4.6 & 0.80 & 0.08 & 2.20 & 0.93 & 0.03 & 2.38 & \underline{0.90} & 0.05 & 2.88 & 0.48 & 0.20 & 2.13 & 0.68 & 0.18 & 2.03 \\
\midrule
\multicolumn{16}{@{}l}{\textit{Trained judge}} \\[2pt]
Trained judge & \underline{0.83} & 0.05 & 3.20 & \underline{0.95} & 0.03 & 2.83 & 0.85 & 0.10 & 2.93 & 0.50 & 0.30 & 2.30 & \underline{0.73} & 0.05 & 2.20 \\
\bottomrule
\end{tabular}
}
\end{table}

\textbf{Settings.} We held the evidence state, the backbone, the retrieval tools and the stop threshold ($\delta = 0.8$) fixed and replaced only the judge $J$ (Section~\ref{sec:judge}). Three stopping rules ignore the state. \emph{Fixed rounds} prompts the agent to use up the four max rounds (we enforced in the prompt but early termination was still up to the agent). \emph{Always continue} always returns \textsc{continue}, leaving the agent to decide when to stop. \emph{Random score} randomly samples stop/continue signal. Three off-the-shelf judges are LLMs prompted zero-shot on the state (Appendix~\ref{app:llm_sufficiency_prompt}): Qwen 3.5~9B, Claude Haiku~4.5 and Claude Sonnet~4.6. The trained judge was fitted once on SciFact-Open and never saw ProClaim-eval. \emph{Iterations at stop} counts the number of times the judge is called, where the max is 4. Every judge was evaluated on the same 40 claims per subset, drawn at random.

\textbf{Results.} \textbf{First}, in Table~\ref{tab:judge}, the trained judge, fitted only on SciFact-Open, has the highest agreement of every stopping rule on SIGNOR, ConnectomeDB and NASA, ahead of the strongest off-the-shelf judge, Sonnet~4.6 (pooled over these three subsets \resMcnJudgeSonnetA{} vs.\ \resMcnJudgeSonnetB{}, \resMcnJudgeSonnetP{}), and the lowest abstention on SIGNOR and NASA (\resJudgeFortyTrainedSignorAbst{} and \resJudgeFortyTrainedNasaAbst{}). It trails Sonnet~4.6 on ExTRI (\resJudgeFortyTrainedExtriAcc{} vs.\ \resJudgeFortySonnetExtriAcc{}) and Haiku~4.5 on NUBASE (\resJudgeFortyTrainedNubaseAcc{} vs.\ \resJudgeFortyHaikuNubaseAcc{}).  \textbf{Second}, the judges that ignore the state are also unstable across subsets. Always continuing and the random score are among the best rules on ExTRI (\resJudgeFortyConstantExtriAcc{} and \resJudgeFortyRandomExtriAcc{}) but the worst on NASA (\resJudgeFortyConstantNasaAcc{} and \resJudgeFortyRandomNasaAcc{}).
\textbf{Third}, among off-the-shelf LLM judges, more capable models are often better judges. Claude Sonnet~4.6 has the highest agreement on four of five subsets, Claude Haiku~4.5 leads on NUBASE (\resJudgeFortyHaikuNubaseAcc{}), and Qwen 3.5~9B has the lowest agreement of any stopping rule on SIGNOR (\resJudgeFortyQwenSignorAcc{}). Every off-the-shelf judge stops after fewer judge calls than the trained judge on every subset; on SIGNOR, Qwen 3.5~9B, Haiku~4.5 and Sonnet~4.6 stop after \resJudgeFortyQwenSignorIter{}, \resJudgeFortyHaikuSignorIter{} and \resJudgeFortySonnetSignorIter{} calls, against \resJudgeFortyTrainedSignorIter{} for the trained judge. In Appendix \ref{sec:ablation1}, we studied this on the full SIGNOR subset, where the off-the-shelf judges score overconfidently and the trained judge is conservative.

\begin{tcolorbox}[
  colback=gray!10,
  colframe=gray!10,
  coltitle=white,
  title={Finding 3: A lightweight trained judge can transfer and replace off-the-shelf LLM judges},
  colbacktitle=black!50!red,
  boxrule=0pt,
  left=10pt,
  right=10pt,
  top=4pt,
  bottom=4pt
]
% n = 40 claims per subset: one claim is 0.025 agreement.
A trained judge that reads the evidence state beats a fixed round budget using fewer steps (pooled agreement \resMcnJudgeFixedA{} vs.\ \resMcnJudgeFixedB{}, \resMcnStmtTwoP{}). A two-layer MLP judge trained only on SciFact-Open, and never exposed to ProClaim-eval, gives the highest agreement of any stopping rule on SIGNOR, ConnectomeDB and NASA, where it matches Sonnet~4.6 (pooled \resMcnJudgeSonnetA{} vs.\ \resMcnJudgeSonnetB{}, \resMcnJudgeSonnetP{}) and outperforms Haiku~4.5 (\resMcnJudgeHaikuA{} vs.\ \resMcnJudgeHaikuB{}, \resMcnStmtOneP{}), hinting at miscalibrated LLM self-evaluation \citep{liu_position_2025}.
\end{tcolorbox}

\section{Related Work}

\paragraph{Scientific claim verification datasets.} 

% Generating claim-level labels requires either large-scale expert adjudication \citep{LoSurdo2025, liu_connectomedb2025_2026} or expensive aggregation of pair-level votes \citep{diggelmann_climate-fever_2020}. 
Prior datasets \citep{diggelmann_climate-fever_2020,saakyan_covid-fact_2021,sarrouti_evidence-based_2021,wadden_fact_2020,wadden_scifact-open_2022} offer verdict labels for claim-passage pairs, which are not suitable for evaluating the consensus claim verification setting. While CIViC-Fact \citep{reisle_evaluating_2025} generated claims from biomedical databases, the distinction is that it also focuses on annotating claim-passage pairs. A separate work \citep{vittor_unifying_2026} validated LLMs for entries in large biomedical knowledge graphs, whereas we focus on recent entries that are difficult for LLMs and cover non-biology domains. Matter-of-Fact \citep{jansen_matter--fact_2025} is a materials science benchmark that offers claim-level labels and aims at open-retrieval verification. However, it focuses on the feasibility of claims. Recent work has highlighted the mismatch between abstract-only evidence and real verification demands. Claims may require compiling evidence from multiple sources \citep{thorne_fever_2018,wadden_scifact-open_2022}, and claims do not reliably appear in paper abstracts \citep{reisle_evaluating_2025,wadden_multivers_2022}. These limitations motivate open-domain, claim-level evaluation for consensus claim verification.

\paragraph{Recursive knowledge construction.} Recent scientific agents \citep{skarlinski_language_2024,mitchener_kosmos_2025,ghareeb_robin_2025} and search-augmented verifiers \citep{wei_long-form_2024,xie_fire_2025,asai_synthesizing_2026} run a loop in which a retrieval step's output becomes the next step's input \citep{zhang_recursive_2025}. The two design choices the loop rests on, what it builds and what controls it, are compared only implicitly. Verifiers and deep-research systems \citep{wei_long-form_2024,xie_fire_2025,asai_synthesizing_2026,huang_deepfact_2026} keep the state in the growing prompt, where retrieved content is used unevenly as the context lengthens \citep{liu_lost_2024}. Scientific agents instead persist structure outside the context, as PaperQA2 does with ranked context state and Kosmos with a world model, but none tests whether that structure provides gains, as we show in Section \ref{sec:exp_state}. The criterion for deciding when to stop the recursion is either a fixed budget \citep{wei_long-form_2024,asai_synthesizing_2026} or the LLM's own self-assessment \citep{asai_self-rag_2024,xie_fire_2025}. As shown by our results in Section \ref{sec:exp_judge}, a trained judge reading the state provides a clear advantage, and the LLM assessment may sometimes be miscalibrated \citep{wen_know_2025,zhou_larger_2024,liu_position_2025}. 

% \paragraph{Harness for LLM agents.}  ReAct \citep{yao_react_2022} established the paradigm of interleaving natural-language reasoning with tool calls, demonstrating that grounded action-taking improves over chain-of-thought alone \citep{wei_chain--thought_2022}. Code-generation agents \citep{chen_program_2023,wang_executable_2024} extended this by executing Python to perform precise arithmetic and symbolic operations. Evidence programming inherits both principles: the agent reasons in natural language via \texttt{nb\_markdown} cells and acts in Python via \texttt{nb\_execute}, producing a verifiable notebook audit trail. Relative to generic agent harnesses, evidence programming is a domain-specific design that provides first-class support for bibliographic operations (PubMed search, NLI-based fact extraction, stance aggregation) as composable, Python functions, and embeds the sufficiency judge as a native stopping oracle rather than an external reward model. DSPy \citep{khattab_dspy_2023} enables compositional LLM programs with declarative optimization, sharing Evidence Programming's goal of structured composability; however, DSPy optimizes prompt selection rather than evidence accumulation. Self-RAG \citep{asai_self-rag_2024} introduces learned retrieval and critique tokens for adaptive stopping in open-domain QA, but relies on a single document context rather than a persistent, mutable evidence state.

\section{Conclusion and Future Work}
\label{sec:conclusion}

% Establishing scientific consensus demands adaptive retrieval, evidence organisation, and reconciliation, a labour-intensive process that does not scale with the pace of literature growth \citep{bachman_automated_2023,wood_empowering_2026}. 

We formalised consensus scientific claim verification, a setting no prior benchmark addresses. For evaluating this verification setting, we introduced a new benchmark, ProClaim-eval, with 718 claims based on five expert-curated databases. Using ProClaim to examine one design axis at a time, we found that the evidence state helps mainly through what it stores. A trained judge that reads the evidence state outperforms fixed-round stopping on every subset, and a lightweight trained judge can transfer and replace off-the-shelf LLM judges. Together, these results support recursively updating a structured evidence state outside the model's context as an effective design for claim verification in scientific agents. For future work, we want to apply this design to curate consensus in scientific databases such as OptimusKG \citep{vittor_unifying_2026} or biological meta-databases such as OmniPath \citep{omnipath}, which could improve the robustness of downstream scientific analyses.

% potentialy to curate consensus on biological meta-databases such as Omnipath \citep{omnipath} or OptimusKG \citep{vittor_unifying_2026} which may improve the robustness of downstream biological analysis.

\section*{Acknowledgments}

JS was supported by the Landesinstitut für Bioinformatikinfrastruktur in Baden-Württemberg. LK acknowledges funding by the German Federal Ministry of Research, Technology, and Space (BMBFTR) with the project 01EJ2501B “Curefib”. YHW acknowledges support from the National Science and Technology Council (NSTC), Taiwan, under grant no. 114-2917-I-006-014.

The costs associated with LLM endpoint usage were supported in the form of API credits, provided by Anthropic through the Anthropic AI for Science program.

\section*{Conflict of Interests}
JSR reports funding from GSK, Pfizer and Sanofi and fees/honoraria from Travere Therapeutics, Stadapharm, Astex, Pfizer, Vera, Grunenthal, Tempus, Moderna and Owkin. AD reports fees from Tempus, MONTAI and Pfizer.

\section*{Reproducibility statement}

We explain how we set up the experiments in Section \ref{sec:experiments}, and provide further details in the Appendix on experimentation and implementation. We also attach the source code for the implementation and experiments as supplementary materials.

\section*{AI disclosure}

In this work, we used generative AI tools to design or provide feedback on research methodology or experiments, to implement methods, to clean and reformat the dataset, and to support qualitative and thematic data analysis and aid in interpreting results. We have not used generative AI tools to generate synthetic data sets, propose or refine hypotheses, and assist with translation. AI uses such as helping develop theoretical models or conceptual frameworks, formulating mathematical claims, providing critical ingredients for proving mathematical claims, and assisting in the writing of proofs do not apply to this work. Additionally, we used generative AI tools to create or modify scientific figures or images, create or edit software code, create artifacts, draft parts of a research paper, brainstorm, sourcing/searching for information, edit a research paper to improve readability, and identify relevant literature. We have reviewed all AI-assisted work. We take responsibility for the final content of this work, including text, claims, or artifacts produced with the aid of generative AI.

\bibliography{iclr2027_conference}
\bibliographystyle{iclr2027_conference}

%%%%%%%%%%%%%%%%%%%%%%%%%%%%%%%%%%%%%%%%%%%%%%%%%%%%%%%%%%%%
\appendix
\newpage
\section*{Table of Contents}
\startcontents[appendix]
\printcontents[appendix]{}{1}{\setcounter{tocdepth}{2}}
%%%%%%%%%%%%%%%%%%%%%%%%%%%%%%%%%%%%%%%%%%%%%%%%%%%%%%%%%%%%
\newpage
%%%%%%%%%%%%%%%%%%%%%%%%%%%%%%%%%%%%%%%%%%%%%%%%%%%%%%%%%%%%
\section{Limitations.}

Results are based on a single frontier model, Claude Sonnet 4.6, at the time when we started our investigation. Owing to the resource constraints, we only ran extensive analyses on a subset of ProClaim-eval in Section \ref{sec:exp_state} and \ref{sec:exp_judge}. ProClaim-eval covers factual claim sentences from three domains instead of free-form claims. While we selected the best suitable databases, the gold labels can inherit the coverage and update lag of the curated databases they come from. Our comparisons focus on one design axis at a time within ProClaim, but ProClaim still differs in prompt and uses several times more compute, and the judge comparison uses 40 claims per subset, so small differences should be read as indicative rather than conclusive. Finally, the trained judge relies on features and training data from biomedicine (SciFact-Open) mainly, and its transfer to other fields is shown only on the NASA and NUBASE subsets.

\section{ProClaim-eval Details.}
\label{app:dataset_creation}

\subsection{Verdict Label Definitions.}
\label{app:verdict_definitions}

Scientific literature, especially in biological and biomedical domains, does not constitute ground truth. Individual papers report findings under specific experimental conditions, and the same phenomenon may be reported differently across studies depending on methodology, statistical power, or the state of knowledge at the time of publication. As a result, verification should not be reduced to a binary decision. Systems restricted to \textsc{support}/\textsc{refute} labels risk systematically misclassifying claims when evidence is inconclusive, internally conflicting, or incomplete.

We adopted a three-way verdict taxonomy motivated by the epistemology of scientific hypothesis testing. In this framework, a claim carries the burden of proof, where the initial position of a claim is non-acceptance, and the evidence must affirmatively establish the claim's validity for a \textsc{support} verdict to be warranted. This is analogous to the logic of null hypothesis significance testing, where the null is maintained until sufficient evidence is gathered to reject it. In particular, we group active contradiction and substantive absence under the \textsc{refute} verdict, which concerns evidential warrant for accepting the claim, not a metaphysical assertion that the claim is false. The \textsc{uncertain} verdict is distinct from \textsc{support} and \textsc{refute} in that relevant evidence exists and engages with the claim, but fails to resolve it. 

% The inclusion of condition (b) for the \textbf{REFUTE} label relies on a closed-world assumption that is fundamental to biological database curation. In an expert-curated knowledge graph, any asserted direct interaction must be explicitly supported by scientific literature. When a thorough search yields no such evidence, it strongly suggests that the original edge assignment was a curation artifact, often stemming from confusable protein aliases or the erroneous conflation of indirect associations with direct interactions. Therefore, the lack of direct evidence warrants edge removal and is classified as \textbf{REFUTE}. The \textbf{UNCERTAIN} label is strictly reserved for cases where direct interaction is claimed by the literature, but the evidence is marginal (e.g., computational simulations without experimental validation) or inherently contradictory.

\begin{tcolorbox}[
  colback=gray!10,
  colframe=gray!10,
  coltitle=white,
  title=Verdict definition,
  colbacktitle=black!50!red,
  boxrule=0pt,
  left=10pt,
  right=10pt,
  top=4pt,
  bottom=4pt
]
We used the same verdict definition (below) in our experimentation for baselines and ProClaim. Minor modifications were made to the LLM-only setting when no retrieval is performed; there would be no external evidence.
\begin{itemize}
    \item \textbf{SUPPORT:} The retrieved evidence contains statements that directly corroborate the claim.
    The evidence, taken at face value, is sufficient to conclude that the claim is true or highly likely true.
    \item \textbf{REFUTE:} Either (a) the retrieved evidence contains statements that directly contradict the claim, or (b) given the scope of the retrieved corpus, a thorough search yields no evidence that substantiates the claim.
    In both cases, the evidence base does not support accepting the claim as true.
    \item \textbf{UNCERTAIN (abstention):} The retrieved evidence is relevant to the claim but is ambiguous, incomplete, or internally conflicting such that neither a clear supportive nor a clear refutatory conclusion can be drawn. This includes cases where evidence partially supports the claim but with meaningful caveats, or where sources of comparable credibility disagree.
\end{itemize}
\end{tcolorbox}

\subsection{Data Generation Process: SIGNOR}
% While the majority of LLM research on knowledge graphs has focused on knowledge graph construction from text via information extraction \citep{pan2024unifying,zhu2024llms}, relatively little attention has been paid to using LLMs to refine existing knowledge graphs by identifying and correcting erroneous edges \citep{huang2025can}. This gap is particularly relevant in biological knowledge curation, where edge correction of gene regulatory networks remains a time-consuming process requiring domain expertise and extensive manual validation. We frame this edge curation task as a claim verification problem, where the claim is defined as the existence of a specific edge in the knowledge graph. For instance, the assertion that BCL2L1 directly activates BAD serves as a verifiable claim. 

We constructed one subset of ProClaim-eval from the SIGNOR database \citep{LoSurdo2025}, a manually curated resource of experimentally supported causal interactions between biological entities. Each claim is expressed as a natural-language statement of the form ``\textit{\{Protein A\} directly activates/inhibits \{Protein B\}}'' This subset comprises two complementary edge sets: a set of ground-truth incorrect edges that should be removed, and a set of ground-truth correct edges that serve as positive controls.

\paragraph{\textsc{Refute} examples.}
We established the set of ground-truth incorrect edges by systematically comparing SIGNOR releases spanning 2018 to 2025. Candidate removals were identified by tracking yearly changes in both SIGNOR IDs and entity names to filter out administrative identifier reassignments. To ensure these removals reflect deliberate curation decisions rather than artifacts of entity deprecation or database restructuring, we retained only those edges whose constituent entities persisted in the October 2025 release.

\paragraph{\textsc{Support} examples.}
To construct a matched set of positive edges, we filtered the October 2025 SIGNOR release (26,127 interactions). We prioritized 3,887 interactions involving proteins from the incorrect edge set, assuming that these entities had undergone more rigorous expert scrutiny. These candidates were ranked by their reference counts in the OmniPath database \citep{omnipath} and passed through a deduplication pipeline that resolved bidirectional redundancies and selected the most-referenced edges for each source and target node. To ensure the benchmark's challenge, we excluded `easy' cases, where the edges are correctly classified by a small language model (gpt-oss-20b) across five independent trials. Following final human expert adjudication, 65 edges were confirmed as the expert-curated base set: 35 \textsc{Support} and 30 \textsc{Refute}.

\paragraph{Claim augmentation via effect flipping.}
To increase the number of \textsc{Refute} examples beyond those derived from historically removed edges, we augmented the dataset by flipping the effect sign of 36 base edges (changing ``activates'' to ``inhibits'' or vice versa).
For 33 of the 35 \textsc{Support} edges, the original edge is confirmed correct, so the sign-flipped claim is labelled \textsc{Refute}.
For 3 \textsc{Refute} edges, the adjudication found the curated effect sign to be wrong, so the sign-flipped claim is labelled \textsc{Support}.
This subset therefore comprises 38 \textsc{Support} and 63 \textsc{Refute}, for a total of 101 claims.

\paragraph{Verdict curation protocol.} We recruited three experts with molecular biology or computational biology background for curation. To ensure the curation process had the same level of standard, we provided human curators with a structured guideline for validating signed directed protein–protein interactions (PPIs) sourced from existing databases.

\begin{enumerate}
    \item \textbf{Claim parsing}: extract source protein, target protein, sign (activation/inhibition), and mechanism.
    \item \textbf{Protein characterisation}: identify functional class, known aliases, and gene family membership using authoritative databases (e.g., UniProt, NCBI Gene).
    \item \textbf{Confusable name check}: flag lexically similar protein names (e.g., CDK2 vs. CK2) that are a major source of annotation error in text-mined databases.
    \item \textbf{Biological plausibility assessment}: evaluate whether the source protein's functional class is consistent with the claimed mechanism and cellular context.
    \item \textbf{Literature evidence search}: conduct systematic searches using gene symbols, aliases, and identified confusable names, examining retrieved papers for both supporting evidence and potential mis-annotation sources (e.g., shared regulators, pathway co-membership).
    \item \textbf{Evidence quality evaluation}: assess experimental methodology, directness, specificity, and sign/direction consistency.
    \item \textbf{Verdict assignment}: classify each interaction with justifications as true, false.
\end{enumerate}

We decided the final verdict based on majority voting over independent expert annotations, where each annotator assigned a stance label to an edge. The majority label was taken as the provisional decision, and cases without a clear majority or with substantive disagreement were escalated to adjudication. During adjudication, experts reviewed the underlying evidence and resolved disagreements to produce a single consensus label.

\subsection{Data Generation Process: ConnectomeDB}

\paragraph{Claim creation.} We included ligand-receptor interactions involved in cell-cell communication based on the ConnectomeDB database. The most recent two releases were in 2020 \citep{hou_predicting_2020} and 2025 (latest) \citep{liu_connectomedb2025_2026}. We constructed this subset by converting curated interaction records from ConnectomeDB into a uniform natural-language claim format ``\textit{\{Ligand\} as ligand directly interacts extracellularly with \{Receptor\} as receptor}'' and for each claim, we extracted the primary ligand and receptor gene symbols. Then, we attached one verdict per claim as described below.

\paragraph{\textsc{Support} examples.}\textsc{Support} examples were created from the newly added interaction set in the 2025 ConnectomeDB release with direct literature support from at least two publications. 

\paragraph{\textsc{Refute} examples.} \textsc{Refute} examples were derived from data in the 2020 release that were removed during the 2025 curation process. We retained only curator-rejected data with specific mechanistic grounds for rejection and excluded entries without primary evidence due to the ambiguity of which evidence was considered. The remaining rejected pairs were labelled deterministically from the curator's comments. Cases such as intracellular interactions, in-cis interactions, wrong directionality, retracted papers, or non-protein interactions were mapped to REFUTE, while cases labelled “old paper without gene symbol” were dropped. Importantly, all examples use the same affirmative claim template.

\paragraph{Dataset.} The final dataset contains 305 claims: 184 \textsc{support}, 121 \textsc{refute}. It combines 184 positive ligand-receptor entries from the ConnectomeDB 2025 release with 121 expert-curation rejected entries from the 2020 release. Each instance includes the claim, label, expert annotator comment, ligand, and receptor symbols and identifiers. For positive examples, provenance is taken from the Perplexity AI summary of the 2025 release curation, which embeds the supporting PubMed identifiers. For rejected examples, provenance includes the curator comment and the PMID of the originally cited publication. The positive subset is supported by literature spanning multiple species, with 145 human, 36 mouse, and 3 rat interactions, while the rejected subset is human-curated and removed in the ConnectomeDB 2025 release. 

\subsection{Data Generation Process: ExTRI}

\paragraph{Claim creation.} We built this subset from the curated sentence corpus behind ExTRI \citep{vazquez_extri_2022}, 47,373 sentences on transcription factor (TF) to target-gene regulation, each labelled independently by at least two domain experts with disagreements resolved to consensus. We did not use the text-mined ExTRI resource, whose edges are machine output. We kept sentences marked valid that passed quality control, excluded the complex identifiers NFKB and AP1, and grouped the remaining 24,256 sentences into 11,354 TF--target pairs. Each claim uses the template ``\textit{\{TF\} directly activates/represses transcription of \{target\}}'', with the direction taken from the curated regulation sign, so the wording does not encode the label.

\paragraph{\textsc{Support} examples.} A pair is \textsc{support} when all of its curated sentences agree on one regulation sign and none negates the regulation. We further required at least five independent supporting papers, so that a \textsc{support} claim reflects accumulated evidence rather than a single report.

\paragraph{\textsc{Refute} examples.} A pair is \textsc{refute} when every curated sentence for it states that the regulation does not occur (a curated negation), with nothing affirming it. The claim states the curated direction affirmatively, and the label follows deterministically from the negation. Curated negations are almost always single-paper, so \textsc{refute} claims are exempt from the five-paper requirement.

\paragraph{Filtering.} To keep claims independent, we selected them so that no paper is shared between two claims (1,504 candidates). We removed `easy' claims that a small language model (gpt-oss-20b) answered correctly in all of five trials, removed a claim whose curated sentences are all non-mammalian, and kept at most two claims per TF for each label.

\paragraph{Dataset.} The final subset contains 125 claims: 49 \textsc{support} and 76 \textsc{refute}. Each instance includes the claim, label, the curated sign or negation, the supporting PubMed identifiers, and the TF and target symbols.

\subsection{Data Generation Process: NUBASE}

\paragraph{Claim creation.} We built this subset from NUBASE2020 \citep{kondev_nubase2020_2021}, the evaluated table of nuclear ground- and isomeric-state properties (5,868 records). A claim asserts that a nuclear isomeric state exists, named by its reported half-life: ``\textit{\{Element\}-\{A\} has an isomeric state with a half-life of about \{value\} \{unit\}}''. Two \textsc{refute} states carry only an upper limit and are phrased ``\dots with a half-life shorter than \{value\} \{unit\}''. Nuclides are written with the element name (e.g.\ Cobalt-52), and units are spelled out.

\paragraph{\textsc{Support} examples.} \textsc{Support} examples are measured isomeric states discovered in 2015 or later. Of 84 candidates, we removed 7 that gpt-oss-20b answered correctly in all of five trials.

\paragraph{\textsc{Refute} examples.} \textsc{Refute} examples are isomeric states that NUBASE2020 flags as reported but non-existent (states claimed in the literature and later disproved). The label follows from this flag. States flagged as of uncertain existence were excluded.

\paragraph{Dataset.} The final subset contains 85 claims: 77 \textsc{support} and 8 \textsc{refute}. Each instance includes the claim, label, the NUBASE2020 status, the nuclide and the state, with the NUBASE2020 publication as evidence.

\subsection{Data Generation Process: NASA Exoplanet Archive}

\paragraph{Claim creation.} We built this subset from the dispositions of the NASA Exoplanet Archive \citep{akeson_nasa_2013,christiansen_nasa_2025}, where every candidate planet carries a curated status. All claims use the template ``\textit{\{object\} is a planet orbiting the star \{star\}}'', and the label follows deterministically from the archive's disposition.

\paragraph{\textsc{Support} examples.} \textsc{Support} examples are planets confirmed in 2024 or later, not flagged as controversial, and backed by at least two distinct publications (excluding the archive's link to its own follow-up service). As for the other subsets, we removed claims that gpt-oss-20b answered correctly in all of five trials.

\paragraph{\textsc{Refute} examples.} \textsc{Refute} examples come from K2 mission objects in two groups. All 8 planets that were published and later refuted, with their refuting papers attached, and 39 of the 242 K2 false positives. We excluded pending candidates, false positives without stated grounds, and false-positive flags from automated vetting pipelines, whose adjudication is not in the citable literature.

\paragraph{Dataset.} The final subset contains 102 claims: 55 \textsc{support} and 47 \textsc{refute}. Each instance includes the claim, label, the archive disposition, the object and its host star, and the supporting references.

\subsection{Data Release}
\label{app:data_release}

To preserve the dataset's value as an evaluation benchmark, we release the full ProClaim-eval in a serialized binary format rather than as plain text. This prevents the dataset from being ingested during web-crawl-based LLM pretraining, which would risk data contamination and undermine its utility for future evaluation. Researchers can load the dataset programmatically using our provided loading scripts. We believe this release strategy balances open access for reproducibility with long-term benchmark integrity.

% \begin{table}[t]
% \centering
% \small
% \caption{Verdict prediction-label agreement, macro false positive rate (FPR), and false negative rate (FNR) on the ProClaim-eval dataset. }
% \vspace{4pt}
% \setlength{\tabcolsep}{2.5pt}
% \renewcommand{\arraystretch}{1.15}

% % Header colors
% \definecolor{headerteal}{RGB}{180,220,220}
% \definecolor{headerlavender}{RGB}{200,190,225}

% % Delta background color scale (green intensity by magnitude)
% \definecolor{deltaxl}{RGB}{144,210,144}   % >=0.30
% \definecolor{deltal}{RGB}{180,225,180}    % 0.20--0.29
% \definecolor{deltam}{RGB}{210,238,210}    % 0.10--0.19
% \definecolor{deltas}{RGB}{232,246,232}    % <0.10

% \input{tables/baseline_metrics_main.tex}
% \label{app:tab:full_results}
% \end{table}

%%%%%%%%%%%%%%%%%%%%%%%%%%%%%%%%%%%%%%%%%%%%%%%%%%%%%%%%%%%%
\section{Experimental details}
\label{app:experiments}

\subsection{Experimental Results in Section \ref{sec:main_results}}
\label{app:main_results_with_f1}

Table~\ref{app:tab:main_results_f1} reports macro-F1 and balanced accuracy for the same runs as Table~\ref{tab:main_results}.

\begin{table}[h]
\centering
\small
\caption{Macro-F1 (F1) and balanced accuracy (BAcc) per ProClaim-eval subset, for the same run behind each cell of Table~\ref{tab:main_results}. \emph{Random} is the mean over its 10 draws. The shaded figure beside each macro-F1 is ProClaim's macro-F1 minus that row's, on the same subset; green where ProClaim is ahead, red where it is behind. The highest value per column is underlined.}
\label{app:tab:main_results_f1}
\vspace{2pt}
\setlength{\tabcolsep}{2.4pt}
\renewcommand{\arraystretch}{1.12}
\resizebox{\linewidth}{!}{\begin{tabular}{@{} l | c c | c c | c c | c c | c c @{}}
\toprule
 & \multicolumn{2}{c}{\cellcolor{headerteal}\textbf{SIGNOR}} & \multicolumn{2}{c}{\cellcolor{headerteal}\textbf{ConnectomeDB}} & \multicolumn{2}{c}{\cellcolor{headerteal}\textbf{ExTRI}} & \multicolumn{2}{c}{\cellcolor{headerteal}\textbf{NUBASE}} & \multicolumn{2}{c}{\cellcolor{headerteal}\textbf{NASA}} \\
\textbf{Method} & F1$\uparrow$ & BAcc$\uparrow$ & F1$\uparrow$ & BAcc$\uparrow$ & F1$\uparrow$ & BAcc$\uparrow$ & F1$\uparrow$ & BAcc$\uparrow$ & F1$\uparrow$ & BAcc$\uparrow$ \\
\midrule
\textit{Random} & 0.38 & 0.32 & 0.40 & 0.33 & 0.38 & 0.32 & 0.32 & 0.34 & 0.37 & 0.31 \\
\midrule
\multicolumn{11}{@{}l}{\textit{LLM Only}} \\[2pt]
Claude Sonnet 4.6 & 0.68\,\gap{gapm}{+0.17} & 0.59 & 0.60\,\gap{gapm}{+0.19} & 0.56 & 0.31\,\gap{gapxl}{+0.47} & 0.25 & 0.05\,\gap{gapxl}{+0.49} & 0.03 & 0.11\,\gap{gapxl}{+0.59} & 0.06 \\
Gemini 2.5 Flash & 0.66\,\gap{gapm}{+0.19} & 0.57 & 0.64\,\gap{gapm}{+0.16} & 0.58 & 0.31\,\gap{gapxl}{+0.47} & 0.28 & 0.28\,\gap{gapl}{+0.25} & 0.39 & 0.32\,\gap{gapxl}{+0.38} & 0.43 \\
\midrule
\multicolumn{11}{@{}l}{\textit{Static Retrieval, S2 Top 5}} \\[2pt]
Claude Sonnet 4.6 & 0.56\,\gap{gapl}{+0.29} & 0.57 & 0.48\,\gap{gapxl}{+0.31} & 0.55 & 0.45\,\gap{gapxl}{+0.33} & 0.51 & 0.16\,\gap{gapxl}{+0.38} & 0.35 & 0.46\,\gap{gapl}{+0.23} & 0.35 \\
Gemini 2.5 Flash & 0.49\,\gap{gapxl}{+0.36} & 0.54 & 0.51\,\gap{gapl}{+0.29} & 0.58 & 0.44\,\gap{gapxl}{+0.34} & 0.52 & 0.18\,\gap{gapxl}{+0.36} & \underline{0.55} & 0.48\,\gap{gapl}{+0.22} & 0.37 \\
\midrule
\multicolumn{11}{@{}l}{\textit{Adaptive Retrieval}} \\[2pt]
ReAct & 0.79\,\gap{gaps}{+0.06} & 0.73 & 0.74\,\gap{gaps}{+0.06} & 0.73 & 0.80\,\gap{gapneg}{-0.02} & 0.73 & 0.16\,\gap{gapxl}{+0.37} & 0.10 & 0.66\,\gap{gaps}{+0.04} & 0.54 \\
FIRE & 0.72\,\gap{gapm}{+0.13} & 0.69 & 0.51\,\gap{gapl}{+0.29} & 0.56 & 0.69\,\gap{gaps}{+0.09} & 0.66 & 0.09\,\gap{gapxl}{+0.45} & 0.33 & 0.72\,\gap{gapneg}{-0.02} & \underline{0.71} \\
OpenScholar & 0.50\,\gap{gapxl}{+0.35} & 0.35 & 0.61\,\gap{gapm}{+0.19} & 0.53 & 0.37\,\gap{gapxl}{+0.41} & 0.25 & 0.03\,\gap{gapxl}{+0.51} & 0.01 & 0.05\,\gap{gapxl}{+0.65} & 0.03 \\
PaperQA2 & 0.69\,\gap{gapm}{+0.16} & 0.65 & 0.74\,\gap{gaps}{+0.06} & 0.73 & 0.83\,\gap{gapneg}{-0.05} & 0.77 & 0.22\,\gap{gapxl}{+0.32} & 0.38 & \underline{0.74}\,\gap{gapneg}{-0.05} & 0.68 \\
Claude Code & 0.82\,\gap{gaps}{+0.03} & 0.79 & 0.76\,\gap{gaps}{+0.04} & 0.75 & \underline{0.86}\,\gap{gapneg}{-0.08} & \underline{0.82} & 0.24\,\gap{gapl}{+0.29} & 0.18 & 0.63\,\gap{gaps}{+0.06} & 0.54 \\
Codex & \underline{0.86}\,\gap{gapneg}{-0.01} & \underline{0.82} & 0.75\,\gap{gaps}{+0.04} & 0.76 & 0.83\,\gap{gapneg}{-0.05} & 0.78 & 0.20\,\gap{gapxl}{+0.34} & 0.16 & 0.64\,\gap{gaps}{+0.05} & 0.56 \\
\midrule
\textbf{ProClaim} & 0.85 & 0.81 & \underline{0.79} & \underline{0.77} & 0.78 & 0.75 & \underline{0.54} & 0.51 & 0.70 & 0.65 \\
\bottomrule
\end{tabular}
}
\end{table}

\subsection{Significance Tests for the Findings}
\label{app:significance}

Each finding statement was tested with a paired comparison of two systems, A and B, on the same claims. The five statements are: (1) the trained judge beats the Haiku~4.5 judge and matches the Sonnet~4.6 judge on SIGNOR, ConnectomeDB and NASA, and (2) the trained judge beats fixed rounds (Finding~3); (3) facts beat abstracts and full text and (4) ProClaim beats in-context accumulation (Finding~2); and (5) ProClaim beats the SOTA harnesses Claude Code and Codex (Finding~1). The unit is a claim scored by both systems. A verdict is correct (1) if it matches the gold label and incorrect (0) otherwise. \textsc{uncertain}, missing and errored answers count as 0. Pooled over the statement's subsets, $b$ counts the claims A gets right and B gets wrong, and $c$ the claims B gets right and A gets wrong. We applied an exact two-sided binomial McNemar test to $b$ against $c$, and a Holm correction across the five statements. A statement that compares A with two systems holds only if A beats both, so its $p$-value is the larger of its two tests (an intersection--union test). Every comparison uses the runs shown in the corresponding table. Statements (2)--(5) pool all five subsets. Statement (1) pools the 40-claim samples of SIGNOR, ConnectomeDB and NASA. Its comparison with Sonnet~4.6 is reported with its two-sided $p$-value and is not part of the Holm correction. Table~\ref{app:tab:mcnemar} reports, for each comparison, the number of claims $n$, the discordant counts $b$ and $c$, the difference $\Delta$ between A's and B's pooled agreement, the exact $p$-value, and the statement's Holm-adjusted $p$-value.

\begin{table}[h]
\centering
\small
\caption{Paired exact McNemar tests of the finding statements, pooled over subsets. $n$: claims; $b$: claims A gets right and B gets wrong; $c$: claims B gets right and A gets wrong; $\Delta$: A's pooled agreement minus B's; $p$: exact two-sided McNemar $p$-value; $p_{\text{Holm}}$: the statement's $p$-value (the larger of its tests), Holm-adjusted across the five statements. }
\label{app:tab:mcnemar}
\vspace{2pt}
\setlength{\tabcolsep}{3pt}
\begin{tabular}{@{} l l r r r r r r @{}}
\toprule
\textbf{Statement: A vs.\ B} & \textbf{Subsets} & $n$ & $b$ & $c$ & $\Delta$ & $p$ & $p_{\text{Holm}}$ \\
\midrule
(1) Trained judge vs.\ Haiku~4.5 judge & 3 $\times$ 40 & 120 & 13 & 2 & +0.092 & 0.0074 & 0.022 \\
Trained judge vs.\ Sonnet~4.6 judge &  & 120 & 12 & 8 & +0.033 & 0.50$^{*}$ &  \\
\midrule
(2) Trained judge vs.\ Fixed rounds & 5 $\times$ 40 & 200 & 35 & 11 & +0.120 & $<$0.001 & 0.0021 \\
\midrule
(3) Facts vs.\ Abstracts & 5 $\times$ 40 & 200 & 30 & 7 & +0.115 & $<$0.001 & 0.024 \\
Facts vs.\ Full text &  & 200 & 24 & 10 & +0.070 & 0.024 &  \\
\midrule
(4) ProClaim vs.\ ReAct + subclaims $-$ state & 5 $\times$ 40 & 200 & 47 & 12 & +0.175 & $<$0.001 & 0.0012 \\
ProClaim vs.\ ReAct + subclaims + facts $-$ state &  & 200 & 42 & 14 & +0.140 & $<$0.001 &  \\
\midrule
(5) ProClaim vs.\ Claude Code & 5, full & 718 & 99 & 65 & +0.047 & 0.0098 & 0.022 \\
ProClaim vs.\ Codex &  & 718 & 104 & 68 & +0.050 & 0.0074 &  \\
\bottomrule
\end{tabular}

\end{table}

% Tables~\ref{app:tab:iclr_budget} and~\ref{app:tab:iclr_evidence_state} give the step budgets behind Table~\ref{tab:main_results} and the construction comparison of Section~\ref{sec:exp_state}. All are generated from the run files by \texttt{scripts/analysis/iclr\_results.py} under the scoring rule of Section~\ref{sec:experiments}.

% \begin{table}[h]
% \centering
% \small
% \caption{Retrieval budget per system and subset. ``Steps'' is the system's own cap on search or tool-calling rounds as recorded in its run configuration; ProClaim stops when its sufficiency judge returns \textsc{stop}, with a safety cap of four rounds.}
% \label{app:tab:iclr_budget}
% \resizebox{\linewidth}{!}{\input{tables/iclr_budget.tex}}
% \end{table}

% \begin{table}[h]
% \centering
% \small
% \caption{Construction procedure at matched retrieval tools (Section~\ref{sec:exp_state}): ReAct given the same paper-search tools as ProClaim, at 10, 20 and 40 steps, against ProClaim. Agreement (Agr.) and abstention rate (Abst.) per subset. ``-'': not yet run.}
% \label{app:tab:iclr_evidence_state}
% \setlength{\tabcolsep}{3pt}
% \input{tables/iclr_evidence_state.tex}
% \end{table}

\subsection{LLM Thinking Mode and Temperature}
All methods, including ProClaim and each baseline, were evaluated with thinking/reasoning mode disabled and a fixed LLM temperature of 0. Disabling the thinking mode ensures that the models operate under comparable inference budgets and that any performance differences could inform the architectural and methodological choices rather than variations in internal chain-of-thought reasoning afforded by extended thinking tokens. Setting the temperature to 0 maximizes the reproducibility of model outputs and comparisons.

\subsection{Computational Resources and Cost}

\begin{table}[t]
\centering
\small
\caption{Mean input (In) and output (Out) tokens per claim, in thousands, for the run behind each cell of Table~\ref{tab:main_results}. Input tokens include cached input. ProClaim counts its main agent only. }
\vspace{4pt}
\setlength{\tabcolsep}{2.5pt}
\renewcommand{\arraystretch}{1.15}

% Header colors
\definecolor{headerteal}{RGB}{180,220,220}
\definecolor{headerlavender}{RGB}{200,190,225}

% Delta background color scale (green intensity by magnitude)
\definecolor{deltaxl}{RGB}{144,210,144}   % >=0.30
\definecolor{deltal}{RGB}{180,225,180}    % 0.20--0.29
\definecolor{deltam}{RGB}{210,238,210}    % 0.10--0.19
\definecolor{deltas}{RGB}{232,246,232}    % <0.10

\begin{tabular}{@{} l | c c | c c | c c | c c | c c @{}}
\toprule
 & \multicolumn{2}{c}{\textbf{SIGNOR}} & \multicolumn{2}{c}{\textbf{ConnectomeDB}} & \multicolumn{2}{c}{\textbf{ExTRI}} & \multicolumn{2}{c}{\textbf{NUBASE}} & \multicolumn{2}{c}{\textbf{NASA}} \\
\textbf{Method} & In & Out & In & Out & In & Out & In & Out & In & Out \\
\midrule
\multicolumn{11}{@{}l}{\textit{LLM Only}} \\[2pt]
Claude Sonnet 4.6 & 0.2 & 0.1 & 0.2 & 0.1 & 0.2 & 0.1 & 0.2 & 0.1 & 0.2 & 0.1 \\
Gemini 2.5 Flash & 0.2 & 0.1 & 0.2 & 0.1 & 0.2 & 0.1 & 0.2 & 0.1 & 0.2 & 0.1 \\
\midrule
\multicolumn{11}{@{}l}{\textit{Static Retrieval, S2 Top 5}} \\[2pt]
Claude Sonnet 4.6 & 2.1 & 0.1 & 2.4 & 0.2 & 2.3 & 0.2 & 1.2 & 0.1 & 1.1 & 0.1 \\
Gemini 2.5 Flash & 1.8 & 0.1 & 2.0 & 0.1 & 1.9 & 0.1 & 1.1 & 0.1 & 1.0 & 0.0 \\
\midrule
\multicolumn{11}{@{}l}{\textit{Adaptive Retrieval}} \\[2pt]
ReAct & 81.8 & 1.5 & 91.4 & 1.7 & 94.8 & 1.5 & 261.2 & 2.3 & 213.3 & 1.6 \\
FIRE & 3.6 & 1.1 & 3.3 & 1.1 & 4.9 & 1.2 & 5.8 & 1.0 & 7.1 & 0.9 \\
OpenScholar & 15.2 & 2.4 & 17.0 & 3.1 & 15.4 & 2.5 & 10.7 & 1.5 & 12.8 & 1.5 \\
PaperQA2 & 218.9 & 29.8 & 236.8 & 32.1 & 219.1 & 30.4 & 528.6 & 50.1 & 163.7 & 19.2 \\
Claude Code & 347.5 & 5.6 & 134.0 & 3.9 & 336.9 & 4.6 & 642.9 & 6.0 & 272.0 & 3.6 \\
Codex & 322.2 & 4.5 & 270.2 & 3.8 & 102.4 & 1.9 & 529.6 & 4.1 & 246.4 & 2.1 \\
\midrule
\textbf{ProClaim} & 626.8 & 13.5 & 612.1 & 14.0 & 721.0 & 14.7 & 1157.9 & 18.3 & 902.1 & 14.8 \\
\bottomrule
\end{tabular}

\label{app:tab:tokens}
\end{table}

All experiments were run on an internal HPC cluster using 8 NVIDIA A100 (80~GB) GPUs, 8~CPU cores, and 64~GB RAM. We served a local Qwen 3.5 9B instance for fact extraction and called the Anthropic API (Claude Sonnet 4.6) as the main agent. For claims from ConnectomeDB (305 cases), total GPU time was $\approx$40 A100 GPU-hours. For claims from SIGNOR (101 cases), the total GPU time was $\approx$12 A100 GPU-hours. For claims from ExTRI (125 cases), NUBASE (85 cases) and NASA (102 cases), the total GPU time was $\approx$13, $\approx$20 and $\approx$17 A100 GPU-hours, respectively. Total GPU time for the main evaluations (Section~\ref{sec:main_results}) was $\approx$102 A100 GPU-hours.

Our architecture follows a modular design where computationally intensive but routine subtasks are handled by a locally hosted model at zero marginal API cost, while the main agent leverages a frontier model for high-level planning. Since the subagent accounts for the majority of LLM calls per claim, hosting it locally on institutional hardware substantially reduces the per-claim cost. The main agent could alternatively use a locally hosted model to reduce API costs. 

The primary temporal bottleneck is the per-paper NLP feature extraction required by the sufficiency judge (Section~\ref{app:sufficiency_classifier_implementation}). For each newly retrieved paper, the pipeline computes named entity coverage (via \texttt{en\_core\_sci\_sm} \citep{neumann2019scispacy}, $\approx$15~MB), semantic similarity (via SBERT \texttt{all-MiniLM-L6-v2} \citep{reimers2019sentence}, $\approx$80~MB), and natural language inference scores (via \texttt{cross-encoder/nli-deberta-v3-large} \citep{he2021debertav3}, $\approx$1.5~GB), which together account for approximately 40\% of total wall time per claim. The main agent (Claude Sonnet 4.6) contributes roughly 25\% of wall time, with an average latency of approximately 30 seconds per agent step. Fact extraction via the local Qwen 3.5 9B subagent accounts for $\approx$10\% of wall time.

Table~\ref{app:tab:tokens} reports the mean input and output tokens per claim for every run in Table~\ref{tab:main_results}. Since ProClaim's main agent re-sends its full execution log (accumulated tool calls and outputs), the input context grows substantially over a claim's trajectory. With prompt caching, most of this repeated prefix is billed at the cached-input rate rather than the full input rate. 

\subsection{Prompts for Compared Methods in Section \ref{sec:main_results}}

We have included updated prompts for all compared methods in Appendix \ref{app:prompts}. When adapting SAFE, FIRE and OpenScholar, we made minimal revisions to the prompts via Claude Sonnet 4.6 for ProClaim-eval's label definitions. 

%%%%%%%%%%%%%%%%%%%%%%%%%%%%%%%%%%%%%%%%%%%%%%%%%%%%%%%%%%%%
\section{ProClaim Details}

\subsection{Implementation}
\label{app:evidence_programming_impl}

We implemented ProClaim in a persistent notebook-style execution environment in which the agent alternates between narrative reasoning and executable evidence updates. The concrete runtime uses domain-specific Python tools for literature search, fact extraction, and stance labelling, while preserving the abstract loop defined in Section~\ref{sec:rkc}.

\paragraph{State representation.} The evidence state is implemented as a structured data model with JSON persistence. In addition to retrieved papers, it stores extracted facts with source identifiers and text spans, detected conflicts between papers, optional subclaim decompositions, and the history of sufficiency scores. Because the state persists after each update, the agent can accumulate evidence across many steps without depending on the context window to remember previously processed papers.

\paragraph{Execution layer.} The primitive operations are realised as Python functions executed in separate kernel sessions. In our current implementation, this includes literature search over PubMed and Semantic Scholar, batched fact extraction and stance assignment, and evidence filtering utilities that reduce neutral or irrelevant papers before sufficiency assessment. Search operations return paper metadata and abstracts. Extraction operations create fact-stance pairs from abstracts or full text, if available. The execution log transcript accumulated so far (markdown reasoning, executed code cells, and their outputs) is sent as user message to the LLM at each step, so the agent can condition its decisions on the full history of evidence operations and their results. 

\paragraph{Programmatic control.} Running inside an executable kernel allows the agent to compose evidence operations with loops, conditionals, and targeted subroutines. This is important when verification requires addressing unresolved subclaims or launching gap-driven retrieval conditioned on the current state. We implemented full paper text access since abstracts may not always be sufficient to resolve a claim \citep{reisle_evaluating_2025,wadden_multivers_2022}. Our implementation is restricted to accessing full texts through licensed APIs (Unpaywall and Elsevier) and does not redistribute available full-text content.

\paragraph{Audit trail.} The notebook also functions as a reproducible trace of the verification process. Executed code cells record evidence-gathering actions and printed outputs, while interleaved markdown cells capture the agent's intermediate reasoning. The saved notebook, therefore, exposes both the final evidence state and the sequence of decisions that produced it.

\paragraph{Caching.} The execution log's length directly determines the size of the input context the model must process. The log is append-only, where each iteration's reasoning, executed code cells, and outputs are appended without modifying or removing earlier entries, so the prompt sent to the agent at step $t$ is always an exact prefix of the prompt at step $t+1$. This stability lets us mark the system prompt and the unchanged prefix of the execution log with cache breakpoints. For Claude Sonnet~4.6, this allows repeated turns to reuse cached prefixes at a 0.1$\times$ read price (with a 1.25$\times$ write surcharge on first use). This significantly reduces per-claim costs without affecting the agent's behaviour or its verdicts.

\paragraph{Hyperparameters.} The hyperparameter $\delta$ gives a threshold of evidence sufficiency for ProClaim to emit a verdict.  For all of our experiments, $\delta$ was set to 0.8. The maximum number of iterations $T_{\max}$ was set to 4 for all experiments. The hyperparameter $\tau$ controls the conflict ratio threshold for sufficiency judge training labels (Section~\ref{app:sufficiency_classifier_implementation}), and it was set to 0.50 for the main experiment.

% Novelty compared to prior work. We include a comparison table (Table \ref{app:tab:novelty_comparison}) against FIRE \citep{xie_fire_2025}, SAFE \citep{wei_long-form_2024} and OpenScholar \citep{asai_synthesizing_2026}, to further illustrate the technical capabilities enabled by ProClaim along evidence storage, retrieval strategy, stopping criteria, and evidence auditability.

% \begin{table}[t]
% \centering
% \caption{Novelty compared to prior work.}
% \label{app:tab:novelty_comparison}
% \begin{tabular}{@{}p{2cm}p{2.5cm}p{2.5cm}p{2.5cm}p{2.5cm}@{}}
% \toprule
% \textbf{System} & \textbf{Evidence storage} & \textbf{Retrieval strategy} & \textbf{Stopping criterion} & \textbf{Evidence auditability} \\
% \midrule
% FIRE &
% LLM prompt context (implicitly compressed as it grows) &
% Fixed multi-round pipeline &
% LLM self-reported confidence &
% Not directly inspectable (embedded in context) \\
% \addlinespace
% SAFE &
% Per-fact independent context (no shared state) &
% Fixed per-fact search loop &
% Fixed iteration budget per fact &
% Not directly inspectable \\
% \addlinespace
% OpenScholar &
% Unstructured text passed to generation &
% Fixed retrieve, generate, refine loop &
% LLM self-feedback &
% Not directly inspectable \\
% \addlinespace
% ProClaim (ours) &
% Structured, persistent state (typed facts, stances, conflicts) &
% Compositional, code-generated (loops/conditionals over evidence state) &
% Trained external MLP classifier &
% Evidence state inspectable verbatim at any point \\
% \bottomrule
% \end{tabular}
% \end{table}

\subsection{ProClaim Tools and Operations}

At the outer loop, ProClaim is given only two generic tools: \texttt{bash} and \texttt{read\_file}. We refer to all functionality exposed to ProClaim as the evidence API, and the Python function schema is injected into the system prompt. This design encourages ProClaim to solve the task through lightweight code generation over a stable, structured interface rather than through a large collection of specialized tools. The \texttt{read\_file} tool complements this setup by allowing ProClaim to inspect persisted artifacts such as \texttt{evidence\_state.json}, \texttt{verdict.json}, and large intermediate outputs when standard output is insufficient.

Below, we describe the main functions available to ProClaim. Each subprocess begins with \texttt{setup\_workspace()}, which initializes the verification session, loads the evidence state, and constructs handles for the LLM subagents. This makes every subprocess idempotent, so ProClaim can resume from disk on every step without depending on hidden conversational state. To retrieve papers, ProClaim uses retrieval functions such as \texttt{search\_pubmed\_llm()} and \texttt{search\_semantic\_scholar()}, which return identifiers for potentially relevant papers. The function \texttt{extract\_and\_add\_facts()} then retrieves available full text and converts it into stance-labelled evidence records stored in the evidence state. Each fact remains linked to its source paper and is stored in a structured format that can be reused across iterations. 

Later iterations can target missing evidence directly through \texttt{search\_for\_gap()}, where ProClaim writes the query. The direct setting also exposes broader corpus expansion functions such as \texttt{search\_semantic\_scholar\_recommendations()}, which let ProClaim move beyond PubMed keyword search via Semantic Scholar Recommendations seeded with papers already judged relevant. Seeds are derived automatically from papers that produced at least one \textsc{support} fact. Papers with only \textsc{refute} facts are used as negative seeds to steer results away from irrelevant directions. Once candidate papers are identified, \texttt{get\_paper\_text()} or \texttt{get\_full\_text\_article()} retrieves the available paper text for downstream extraction.

Before deciding whether it has enough evidence, ProClaim calls \texttt{populate\_paper\_features()}. This function computes the natural language and metadata-derived feature representation consumed by the sufficiency judge. The sufficiency decision itself is made by \texttt{check\_sufficiency()} to return a confidence score. In addition, this automatically invokes LLM-based gap findings from the current evidence state. When the accumulated evidence is sufficient, ProClaim finalizes the run
with \texttt{emit\_verdict()}, which writes a structured verdict object containing the final label, reasoning, and supporting evidence references. If the evidence remains insufficient, ProClaim
continues the loop by reading the latest sufficiency output or sufficiency history, issuing new gap-targeted searches, and repeating the extraction and feature-population steps. 

\subsection{Dataset for training the sufficiency judge}
\label{app:sufficiency_dataset}

We intended to label a state \textsc{sufficient} when the evidence is very clear and unanimous, while any conflicting support or refute findings or added noise makes it \textsc{insufficient}. That means \textsc{sufficient} here does not mean there is strong evidence, but that the current evidence state is a clean, internally consistent stopping state. So the model was trained to withhold a positive label unless the whole pool looks settled.

We trained the MLP on the SciFact-Open dataset, which contains 279 claims with labels assigned at the claim-paper pair level. Of these, 206 claims have at least one evidence paper labelled as either \textsc{support} or \textsc{refute}, and 73 claims have no supporting evidence. To quantify the degree of conflict between supporting and refuting evidence, we define the minority ratio as $r_{\text{minority}} = \min(N_{\text{support}}, N_{\text{refute}}) / (N_{\text{support}} + N_{\text{refute}})$ when at least one paper provides a stance (i.e., $N_{\text{support}} + N_{\text{refute}} > 0$). If $r_{\text{minority}} \geq \tau$,  the state is labelled insufficient ($y=0$) and contains conflicting support/contradict evidence or irrelevant noise papers sampled for claims without evidence annotations. For the 73 NEI claims, we performed data augmentation by randomly appending 1 to 3 irrelevant papers and labelling the resulting states as insufficient to train the model's ability to recognise irrelevant evidence. Otherwise, the state is categorized as sufficient ($y=1$). In Section \ref{sec:judge}, we used $\tau=0.5$ as a simple criterion to decide conflicting states as they have equal amount of supporting and refuting evidence.

For each claim-paper pair, we extracted metadata features from PubMed/OpenAlex, including publication year, journal impact proxy, normalized citation count, and maximum author $h$-index, together with NLP features computed from the paper full text when available, otherwise from the abstract. The NLP features are claim--evidence semantic similarity, biomedical entity coverage, and NLI entailment/contradiction/neutral scores, computed over overlapping text chunks for long documents.

\subsection{Sufficiency judge Implementation}
\label{app:sufficiency_classifier_implementation}

For each paper $P_i \in \mathcal{E}$, we extracted a feature vector $\mathbf{v}_i = \psi(\mathcal{C}, P_i) = [\mathbf{v}_i^{\text{meta}}, \mathbf{v}_i^{\text{nlp}}] \in \mathbb{R}^k$ combining publication quality metadata with NLP-based signals. These local vectors are aggregated into a global representation $\mathbf{v}_{\text{global}} = g(\mathbf{v}_1, \ldots, \mathbf{v}_N) \in \mathbb{R}^d$ via pooling and cross-weighting statistics that summarize corpus-level consensus and controversy. These paper-level signals are aggregated into a 10-dimensional state representation.

We trained a two-hidden-layer MLP with hidden dimensions $64$ and $32$ using class-weighted binary cross-entropy to predict the sufficiency label,
\[
\hat{y} = f_{\phi}(\mathcal{C}, \mathcal{E}) = \sigma\!\left(\mathrm{MLP}(\mathbf{v}_{\mathrm{global}})\right),
\]
where $\mathbf{v}_{\mathrm{global}}$ denotes the aggregated feature representation of the current evidence state for claim $\mathcal{C}$. We used a stratified $70/15/15$ train/validation/test split. At inference time, the model outputs a sufficiency score for the full evidence state, which serves as the stopping criterion in the recursive retrieval loop.

The sufficiency MLP classifier was trained to reward clean, consensus-like states. Its input is a 10-dimensional vector pooled over the papers in the state: (1) mean claim--paper similarity, (2) mean entity coverage, (3) contradiction ratio (the fraction of papers whose most likely NLI class is contradiction), (4) controversy index (the entropy of the papers' NLI stance votes), (5) recency-weighted entailment ($\sum_i e^{-0.1\,a_i}\, p^{\text{ent}}_i$, with $a_i$ the paper's age in years), (6) mean normalised citation count, (7) mean log impact factor, (8) year span, (9) number of papers, and (10) number of full texts. To ensure reliability of the sufficiency signal, a sufficient label is forced back to insufficient when the state holds fewer than 3 papers, except in the last round. This fallback applies to every judge that can return sufficient, including the off-the-shelf LLM judges and the random score in Section~\ref{sec:exp_judge}. This design aims at making the whole system keep searching unless the evidence state is both strong and structurally clean. From our experiments, we found that the sufficiency labels are not monotonic. Adding papers can lower consensus features or increase controversy, so a run can become sufficient and then fall back to insufficient.

% \subsection{Ablations}
% \label{app:ablations}

% The judge ablation (Section~\ref{sec:ablation1}) was conducted exclusively on the SIGNOR subset. Total GPU time is approximately 300 A100 GPU-hours. We prioritise the ablation on SIGNOR over ConnectomeDB because SIGNOR claims have been re-curated internally and therefore provided us with a more controlled environment.

\subsection{Judge Ablation: How Does the Choice of Judge Affect ProClaim's Performance?}  
\label{sec:ablation1}
\begin{figure}[t]
    \centering
    \includegraphics[width=\linewidth]{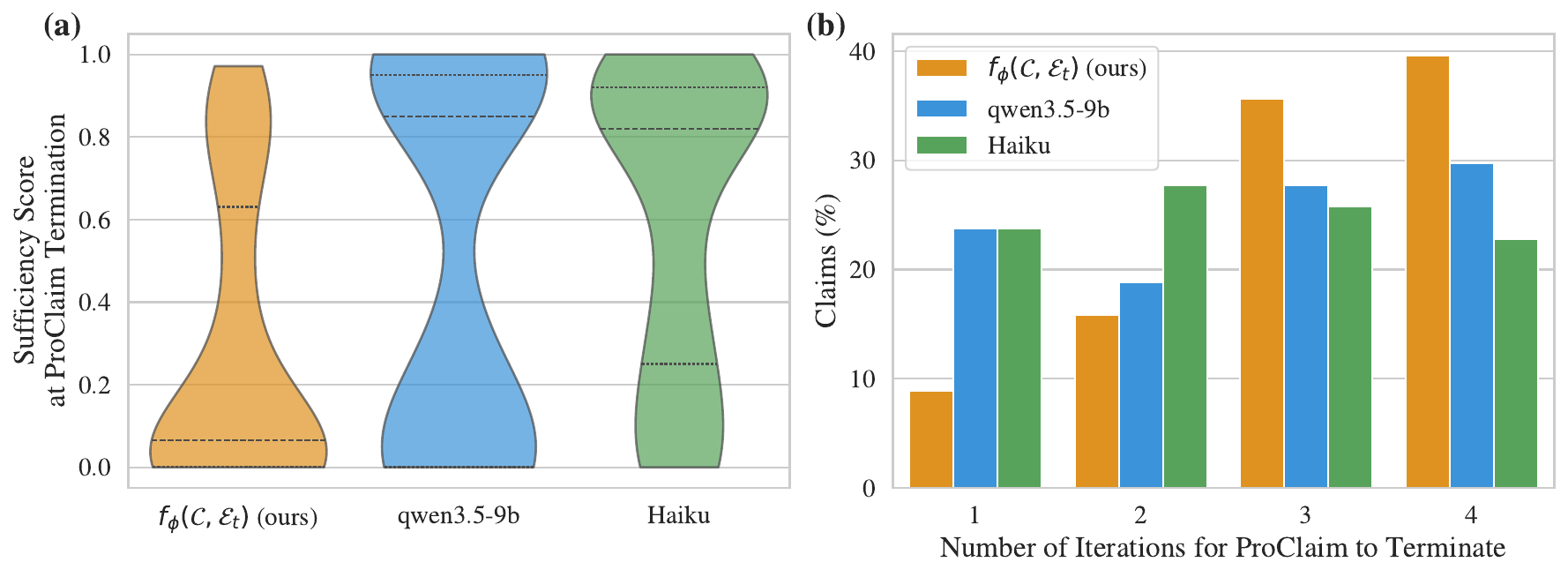}
    \caption{%
         Effect of the judge on sufficiency scoring and ProClaim termination. \textbf{(a)}~The trained judge is conservative, assigning low scores for most claims and driving the agent to collect more evidence across multiple rounds. Qwen 3.5 9B exhibits a bimodal distribution, with a dominant mode near 1.0 and a secondary mode near 0.0. Haiku assigns uniformly high scores with almost no mass near zero. \textbf{(b)}~The trained judge's conservative scoring terminates ProClaim at three and four iterations, while Qwen 3.5 9B's bimodal pattern produces peaks at both one and four iterations, and Haiku distributes more evenly.
    }
    \label{fig:sufficiency-backends}
\end{figure}

We compared the trained judge $J(c, s_t)$ with two off-the-shelf LLM judges, Qwen 3.5~9B and Claude Haiku~4.5, prompted zero-shot on the same evidence state (Appendix~\ref{app:llm_sufficiency_prompt}), on the full SIGNOR subset. All other settings follow Section~\ref{sec:exp_judge}. Agreement and abstention were computed as in the main text, and precision (P), recall (R) and F1 were computed for the two gold labels. Table~\ref{tab:agent-performance} shows that the trained judge has the highest agreement (\resAblJudgeTrainedAcc{} vs.\ \resAblJudgeQwenAcc{} for Qwen 3.5~9B and \resAblJudgeHaikuAcc{} for Haiku~4.5) and macro F1 (\resAblJudgeTrainedMacroFone{} vs.\ \resAblJudgeQwenMacroFone{} and \resAblJudgeHaikuMacroFone{}), with the highest \textsc{refute} recall (\resAblJudgeTrainedRefR{} vs.\ \resAblJudgeQwenRefR{} and \resAblJudgeHaikuRefR{}), while Haiku~4.5 has the highest \textsc{support} recall (\resAblJudgeHaikuSupR{}). The trained judge also keeps the recursion running longest (\resAblJudgeTrainedIter{} judge calls at stop vs.\ \resAblJudgeQwenIter{} and \resAblJudgeHaikuIter{}). Both LLM judges assign higher sufficiency scores than the trained judge on average (Figure~\ref{fig:sufficiency-backends}a). Haiku~4.5's scores cluster near 1.0, the systematic overconfidence discussed in prior work \citep{wen_know_2025,zhou_larger_2024}, and Qwen 3.5~9B's are bimodal, near one or near zero for most claims. The trained judge is fitted on SciFact-Open, which covers general biomedical topics, and is used on SIGNOR without domain-specific fine-tuning.

% \begin{tcolorbox}[
%   colback=gray!10,
%   colframe=gray!10,
%   coltitle=white,
%   title=Result 2,
%   colbacktitle=black!50!red,
%   boxrule=0pt,
%   left=10pt,
%   right=10pt,
%   top=4pt,
%   bottom=4pt
% ]
% Table~\ref{tab:agent-performance} shows that the trained MLP classifier ($f_\phi$) achieves the highest overall agreement (0.75) and macro F1 (0.58), outperforming Qwen 3.5 9B (0.70/0.57) and Haiku (0.72/0.52). This is accompanied by the longest retrieval loops (mean 3.06 iterations vs.\ 2.63/2.48), indicating that our trained sufficiency judge invokes more evidence retrieval iterations. 
% \end{tcolorbox}

\begin{table}[t]
\centering
\caption{Judge ablation on SIGNOR (101 claims), with the evidence state, backbone, retrieval tools and stop threshold held fixed. Agreement (Agr., abstention counted as a disagreement), abstention rate (Abst.), macro F1 over the two gold labels, mean judge calls at stop (Iter.), and precision (P), recall (R) and F1 per gold label. }
\label{tab:agent-performance}
\setlength{\tabcolsep}{3pt}
\begin{tabular}{@{} l c c c c | c c c | c c c @{}}
\toprule
 & & & & & \multicolumn{3}{c|}{\textsc{support}} & \multicolumn{3}{c}{\textsc{refute}} \\
 & Agr.$\uparrow$ & Abst. & Macro F1$\uparrow$ & Iter. & P & R & F1 & P & R & F1 \\
\midrule
Trained judge & \underline{0.80} & 0.12 & \underline{0.85} & 3.21 & 0.82 & 0.84 & 0.83 & 0.98 & 0.78 & 0.87 \\
Qwen3.5 9B & 0.71 & 0.18 & 0.78 & 2.63 & 0.78 & 0.82 & 0.79 & 0.95 & 0.65 & 0.77 \\
Claude Haiku 4.5 & 0.76 & 0.11 & 0.81 & 2.48 & 0.74 & 0.92 & 0.82 & 0.98 & 0.67 & 0.79 \\
\bottomrule
\end{tabular}

\end{table}

\subsection{LLM Judge Prompt}
\label{app:llm_sufficiency_prompt}

To evaluate the impact of replacing our MLP sufficiency judge with an LLM, we designed an ablation study using an LLM-based sufficiency judge. The LLM is prompted to act as a scientific evaluator. It assesses whether the provided reference context contains sufficient evidence to support or refute the claim. The complete prompt template used for this LLM-based evaluator is presented below.

\begin{tcolorbox}[breakable, colback=gray!5,colframe=blue!70,title=Prompt Template: LLM sufficiency judge, fonttitle=\bfseries]
\small
\noindent You are a scientific evaluator.
Given a CLAIM and a REFERENCE (paper titles, abstracts, and NLP features), assess whether the REFERENCE contains sufficient evidence to support or refute the CLAIM.

\vspace{0.5em}
\noindent Assign a sufficiency score between 0.0 and 1.0:
\begin{itemize}
    \item \textbf{1.0}: The REFERENCE contains decisive evidence --- either clearly supporting OR clearly refuting the CLAIM.
    Both directions are equally "sufficient".
    Papers that unanimously refute the claim deserve the same high score as papers that unanimously support it.
    \item \textbf{0.0}: The REFERENCE is entirely irrelevant (entities, topic, and semantics are unrelated to the claim), or no papers were found.
    \item Values in between reflect partial relevance, inconclusive evidence, or genuinely conflicting evidence.
    Conflicting evidence means the papers are relevant and credible, but split in direction --- after weighing their content, methodology, and credibility, no clear verdict can be reached.
    Do NOT judge conflict by the ratio of papers alone; a single high-quality paper can outweigh several weaker ones.
\end{itemize}

\vspace{0.5em}
\noindent \textbf{\#\#\# FEATURE DEFINITIONS} \\
Each paper in the REFERENCE includes pre-computed Metadata and NLP fields:

\vspace{0.5em}
\noindent \textbf{Metadata:}
\begin{itemize}
    \item \texttt{publication\_year}: Year the paper was published.
    \item \texttt{log\_impact\_factor}: log(1 + journal impact factor).
    Higher = more prestigious venue.
    \item \texttt{normalized\_citation\_count}: Citations / paper age.
    Measures community attention.
    \item \texttt{author\_h\_index\_max}: Highest H-index among all authors.
    Proxy for author credibility.
\end{itemize}

\vspace{0.5em}
\noindent \textbf{NLP (computed relative to the CLAIM):}
\begin{itemize}
    \item \texttt{claim\_entity\_coverage}: Fraction of claim biomedical entities also in the paper (0--1).
    \item \texttt{semantic\_similarity}: SBERT cosine similarity between claim and abstract (0--1).
    \item \texttt{nli\_entailment}: DeBERTa NLI probability that the \texttt{nli\_best\_chunk\_text} ENTAILS the claim (0--1).
    \item \texttt{nli\_contradiction}: Probability the \texttt{nli\_best\_chunk\_text} CONTRADICTS (refutes) the claim (0--1).
    \item \texttt{nli\_neutral}: Probability the \texttt{nli\_best\_chunk\_text} is NEUTRAL (0--1).
    Sum $\approx$ 1.
    \item \texttt{nli\_best\_chunk\_text}: The most relevant text passage used to compute NLI scores.
\end{itemize}

\vspace{0.5em}
\noindent \textbf{\#\#\# CLAIM} \\
\{claim\}

\vspace{0.5em}
\noindent \textbf{\#\#\# REFERENCE} \\
\{reference\}

\vspace{0.5em}
\noindent First, output a brief \textbf{\#\#\# EXPLANATION} (plain text reasoning). \\
Then output \textbf{\#\#\# EVALUATION} as a JSON object with exactly these keys:

\vspace{0.5em}
\noindent \texttt{\{} \\
\texttt{\ \ "sufficiency\_score": \textless float 0.0--1.0\textgreater,} \\
\texttt{\ \ "reasoning": "\textless one-sentence summary\textgreater"} \\
\texttt{\}}

\vspace{0.5em}
\noindent Output ONLY the EXPLANATION block followed by the EVALUATION JSON block.
No other text.
\end{tcolorbox}

\subsection{ProClaim Prompts}

\begin{tcolorbox}[breakable, colback=gray!5,colframe=blue!70,title=ProClaim System Prompt, fonttitle=\bfseries]
\footnotesize
\begin{Verbatim}[breaklines,breakanywhere,showtabs=false,breaksymbol=]
You are an evidence-programming agent that verifies scientific claims and produces verdicts [{verdict_names}].
{verdict_definitions}

You work by writing Python code executed via the `bash` tool.  The `command`
argument you pass to `bash` must be the literal shell command
`python3 -c '...'` with your Python code inside the quotes — NOT the bare
Python code by itself (bare Python code passed directly as the shell command
will fail with `command not found` / syntax errors, since bash does not know
Python syntax).  All evidence API functions are available as Python imports.

## First step — set up

Call `bash` with the setup code to bootstrap the workspace. The `command`
argument must look exactly like this (a single shell string, `python3 -c`
followed by your code in quotes):

```bash
python3 -c '
from proclaim.verification.evidence_api import setup_workspace
state, llm, workspace = setup_workspace(
    claim="{claim}",
    workspace_path="{workspace}",
)
print("Ready")
'
```

After this, every subsequent bash call must re-load state from disk
(there is no persistent kernel).  Use setup_workspace which is idempotent
(loads existing state if present). Again, wrap it in `python3 -c '...'`:

```bash
python3 -c '
from proclaim.verification.evidence_api import setup_workspace
state, llm, workspace = setup_workspace(claim="{claim}", workspace_path="{workspace}")
# ... your evidence API calls here ...
'
```

{function_docs}

{schemas}

## Workflow

1. Call bash with the setup code above.
2. Decompose the claim into 1–5 atomic subclaims, each a single independently
   verifiable assertion. Use 1 (the original claim) if the claim is already simple
   enough to search directly. Include alternative names or aliases for key entities.
   Set subclaims before searching: `state.subclaims = [...]` then `state._auto_save()`.
{subclaim_examples}
3. Search (iteration 0):
   a. call search_pubmed_llm(state.claim, state, llm) — runs two LLM-generated PubMed queries (claim-only and subclaim-enriched), deduplicated
   b. call search_semantic_scholar_dual(state.claim, state, llm) — runs two S2 queries (claim-only and subclaim-enriched), covers bioRxiv preprints and non-MEDLINE journals
4. Extract facts: call extract_and_add_facts(llm, pmids, state, max_workers=8)
  to process all newly retrieved papers in parallel.  Do NOT loop over paper IDs.
5. Call populate_paper_features(state) — REQUIRED before check_sufficiency().
6. Call filter_papers_by_stance(state) — removes papers with only neutral facts.
7. Check sufficiency: result = check_sufficiency(state, llm); print(result)
8. Call get_sufficiency_history(state) to monitor the confidence trend.
9. If insufficient: use gap-targeted retrieval:
   a. search_for_gap(gap_description, state)
   b. search_semantic_scholar_recommendations(state) — S2 graph expansion
   c. formulate_gap_queries(llm, state) — LLM-generated gap queries
10. Repeat until confidence >= {sufficiency_threshold} or {max_iterations} iterations.{sufficiency_mode_note}
11. Call emit_verdict(...) to produce the final verdict.

## Rules

- Use bash for ALL evidence API calls — write Python code.
- Print results to stdout so you can see them.
- ALL state mutations are auto-saved to evidence_state.json.
- `state` does NOT persist across bash calls — reload it each time
  (or call setup_workspace again).
- When emit_verdict is called, the verification is complete.
- Do NOT attempt to debug, patch, or work around evidence API functions that return 0 facts. A 0-fact result means the paper lacks relevant evidence or accessible full text — not a tool bug. Move on to other papers or emit a verdict.

## CRITICAL: Grounded Evidence Only

- NEVER fabricate facts.  Every fact must come from a retrieved paper.
- Use extract_and_add_facts(llm, pmids, state) — pass the full paper ID list.
- If extraction returns 0 for multiple paper IDs, use refine_search_for_failed_papers:
  failed_ids = [paper_id for paper_id, count in results.items() if count == 0]
  new_ids = refine_search_for_failed_papers(failed_ids, state, llm, max_new_papers=5)
  results2 = extract_and_add_facts(llm, new_ids, state)
- NEVER call add_facts_from_dicts with manually written text.
- If no papers contain relevant evidence, say so in the verdict.
- add_extraction_context_note is for SYNONYM/ALIAS MAPPINGS ONLY.  Never write 
  search goals, task descriptions, or paper-specific findings into it.

## Feature Computation for MLP Classifier

You MUST call populate_paper_features(state) after extracting facts and
before check_sufficiency().  Without it, the MLP classifier receives
all-zero features and predictions will be inaccurate.

Required sequence in EVERY iteration:
1. search (PubMed + Semantic Scholar)
2. extract_and_add_facts(llm, pmids, state)
3. populate_paper_features(state)           ← REQUIRED
4. filter_papers_by_stance(state)           ← REQUIRED (removes neutral/irrelevant papers)
5. check_sufficiency(state, llm)

## Important

- Workspace path: {workspace}
- Max iterations: {max_iterations}
- Sufficiency threshold: {sufficiency_threshold}

## Common Python Mistakes to Avoid

- When building the `reasoning` string for emit_verdict, compute all values
  BEFORE constructing the string.  Use f-strings (prefix with `f`) or
  `.format()`, never bare curly braces in a regular string.
  BAD:  reasoning = "Found {{len(facts)}} facts"   — literal braces, not evaluated
  GOOD: n = len(facts); reasoning = f"Found {{n}} facts"  — f-string evaluates n
\end{Verbatim}
\end{tcolorbox}

\begin{tcolorbox}[breakable, colback=gray!5,colframe=blue!70,title=ProClaim Fact Extraction, fonttitle=\bfseries]
\footnotesize
\begin{Verbatim}[breaklines,breakanywhere,showtabs=false,breaksymbol=]
You are a scientific fact extraction specialist.

Given a paper and a claim with subclaims, extract every atomic fact that addresses the claim, accounting for varying terminology, synonyms, or aliases used in the text.

For each fact provide a JSON object with:
- "text": factual statement (one sentence, self-contained)
- "stance": one of {stance_options}
- "source_pmid": "{source_pmid}"
- "confidence": 0.0-1.0, how clearly the paper states this

Example: [{{"text": "Gs alpha directly stimulates adenylyl cyclase activity.", "stance": "SUPPORT", "source_pmid": "{source_pmid}", "confidence": 0.9}}]

Stance definitions:
{stance_block}

Rules:
- Base facts strictly on the provided text. Recognize equivalent terms, but do not hallucinate logical leaps not present in the paper.
- If the paper has no relevance to the claim or subclaims, return [].
- Each fact must be independently verifiable from the source paper.
Now extract facts for the following:

Claim: {claim}

Subclaims:
{subclaims_str}
{context_block}
Paper (paper ID: {source_pmid}):
{paper_text}

Output ONLY a JSON array of fact objects. No other text.
\end{Verbatim}
\end{tcolorbox}

\begin{tcolorbox}[breakable, colback=gray!5,colframe=blue!70,title=ProClaim PubMed Query Generation, fonttitle=\bfseries]
\footnotesize
\begin{Verbatim}[breaklines,breakanywhere,showtabs=false,breaksymbol=]
You are a biomedical search expert. Generate a comprehensive PubMed query for the following scientific claim:

Claim: {claim}

Requirements:
- Include key entities (genes, mutations, diseases, drugs, proteins)
- Add synonyms and common variants using OR operators where appropriate
- Use appropriate field tags when beneficial (e.g., [Title/Abstract], [MeSH])
- Keep query length reasonable (5-20 words)
- Use proper PubMed syntax with Boolean operators (AND, OR, NOT)
- Use parentheses to group related terms

Examples:
- Diagnosis claim: "JAK2 V617F is not associated with lymphoid leukemia"
  → (JAK2 AND V617F) AND (lymphoid leukemia OR B-ALL OR T-ALL OR CLL)

- Drug resistance claim: "NT5C2 K359Q mutation does not confer resistance to Nelarabine"
  → (NT5C2 AND K359Q) AND (Nelarabine OR Arabinosylguanine) AND (resistance OR sensitivity)

- PPI claim: "MAPK1 activates H3-3A through phosphorylation"
  → (MAPK1 OR ERK2) AND (H3-3A OR HIST1H3A) AND (phosphorylation OR activation)

Output ONLY the query string without any explanation or markdown formatting.
\end{Verbatim}
\end{tcolorbox}

\begin{tcolorbox}[breakable, colback=gray!5,colframe=blue!70,title=ProClaim Semantic Scholar Query Generation, fonttitle=\bfseries]
\footnotesize
\begin{Verbatim}[breaklines,breakanywhere,showtabs=false,breaksymbol=]
You are a biomedical search expert. Generate a Semantic Scholar query for the following scientific claim:

Claim: {claim}{subclaims_section}

Requirements:
- Use plain keyword phrases (Semantic Scholar does not support PubMed field tags)
- Include alternative names, gene symbols, aliases, and common synonyms for key entities
- Keep query concise (3-15 words)
- Prefer terms that appear in titles and abstracts of relevant papers

Examples:
- PPI claim: "MAPK1 activates H3-3A through phosphorylation"
  → MAPK1 ERK2 H3.3 phosphorylation activation

- Drug resistance: "NT5C2 K359Q mutation does not confer resistance to Nelarabine"
  → NT5C2 K359Q nelarabine arabinosylguanine resistance

Output ONLY the query string without any explanation or markdown formatting.
\end{Verbatim}
\end{tcolorbox}

\begin{tcolorbox}[breakable, colback=gray!5,colframe=blue!70,title=ProClaim Gap-Targeted Query Generation, fonttitle=\bfseries]
\footnotesize
\begin{Verbatim}[breaklines,breakanywhere,showtabs=false,breaksymbol=]
Generate a focused PubMed query to find evidence for this gap:

Claim: {claim}
Gap: {gap_description}

Generate a query that specifically targets this missing evidence. Keep it concise (3-10 words).
Output ONLY the query string without explanation.
\end{Verbatim}
\end{tcolorbox}

\section{Prompts for Baselines and Compared Methods}
\label{app:prompts}

Static retrieval searches S2 with the claim text. On NUBASE it searches the nuclide followed by ``isomer half-life'' (e.g.\ ``Cobalt-52 isomer half-life''), because the full claim sentence returns no paper for 83 of the 85 claims.

\begin{tcolorbox}[breakable, colback=gray!5,colframe=blue!70,title=Retrieval Final Verdict System Prompt, fonttitle=\bfseries]
\footnotesize
\begin{Verbatim}[breaklines,breakanywhere,showtabs=false,breaksymbol=]
You are a scientific claim verification expert.

Given a claim and retrieved evidence passages, assign one of the following verdicts:

- SUPPORT — The retrieved evidence contains statements that directly corroborate the claim. The evidence, taken at face value, is sufficient to conclude that the claim is true or highly likely true.
- REFUTE — Either (a) the retrieved evidence contains statements that directly contradict the claim, or (b) given the scope of the retrieved corpus, a thorough search yields no evidence that substantiates the claim. In both cases, the evidence base does not support accepting the claim as true.
- UNCERTAIN — The retrieved evidence is relevant to the claim but is ambiguous, incomplete, or internally conflicting such that neither a clear supportive nor a clear refutatory conclusion can be drawn. This includes cases where evidence partially supports the claim but with meaningful caveats, or where sources of comparable credibility disagree.

Respond with valid JSON only — no markdown fences, no extra keys:
{
    "label": "SUPPORT" | "REFUTE" | "UNCERTAIN",
    "reasoning": "One or two sentences citing specific evidence",
    "evidence": ["PMID1", "PMID2"]
}
\end{Verbatim}
\end{tcolorbox}

\begin{tcolorbox}[breakable, colback=gray!5,colframe=blue!70,title=Retrieval Final Verdict User Template, fonttitle=\bfseries]
\footnotesize
\begin{Verbatim}[breaklines,breakanywhere,showtabs=false,breaksymbol=]
Claim: {claim}

Retrieved Evidence:
{evidence}

Based on the above evidence, classify the claim. Output JSON.
\end{Verbatim}
\end{tcolorbox}

\begin{tcolorbox}[breakable, colback=gray!5,colframe=blue!70,title=LLM-Only Final Verdict System Prompt, fonttitle=\bfseries]
\footnotesize
\begin{Verbatim}[breaklines,breakanywhere,showtabs=false,breaksymbol=]
You are a scientific claim verification expert.

Given a claim, assess whether it is supported by established scientific knowledge:

- SUPPORT — Your knowledge of the scientific literature supports the claim as true or highly likely true based on established experimental evidence.
- REFUTE — Your knowledge of the scientific literature contradicts the claim, or the claim is not substantiated by any known experimental evidence.
- UNCERTAIN — The scientific literature is ambiguous, incomplete, or conflicting on this claim, or your knowledge is insufficient to make a reliable determination.

Respond with valid JSON only — no markdown fences, no extra keys:
{
    "label": "SUPPORT" | "REFUTE" | "UNCERTAIN",
    "reasoning": "One or two sentences explaining your verdict"
}
\end{Verbatim}
\end{tcolorbox}

\begin{tcolorbox}[breakable, colback=gray!5,colframe=blue!70,title=LLM-Only Final Verdict User Template, fonttitle=\bfseries]
\footnotesize
\begin{Verbatim}[breaklines,breakanywhere,showtabs=false,breaksymbol=]
Claim: {claim}

Based on your scientific knowledge, classify the claim. Output JSON.
\end{Verbatim}
\end{tcolorbox}

\begin{tcolorbox}[breakable, colback=gray!5,colframe=blue!70,title=ReAct System Prompt (Semantic Scholar), fonttitle=\bfseries]
\footnotesize
\begin{Verbatim}[breaklines,breakanywhere,showtabs=false,breaksymbol=]
You are a scientific claim verification agent using the ReAct framework.

Your task: determine whether a scientific claim is "SUPPORT" or "REFUTE" or "UNCERTAIN" based on evidence you retrieve from academic literature.

Verdict definitions:
   - "SUPPORT" — The retrieved evidence contains statements that directly corroborate the claim. The evidence, taken at face value, is sufficient to conclude that the claim is true or highly likely true.
   - "REFUTE" — Either (a) the retrieved evidence contains statements that directly contradict the claim, or (b) given the scope of the retrieved corpus, a thorough search yields no evidence that substantiates the claim. In both cases, the evidence base does not support accepting the claim as true.
   - "UNCERTAIN" — The retrieved evidence is relevant to the claim but is ambiguous, incomplete, or internally conflicting such that neither a clear supportive nor a clear refutatory conclusion can be drawn. This includes cases where evidence partially supports the claim but with meaningful caveats, or where sources of comparable credibility disagree.

Available tools:
- search_papers(query): Search Semantic Scholar for academic papers. Returns titles, abstracts, and metadata. Use specific queries targeting the entities and relationships in the claim.

Strategy:
1. Think step-by-step about what evidence you need.
2. Issue targeted search queries to find that evidence.
3. After each search result, assess whether you have enough evidence.
4. When confident, respond with your final verdict in the EXACT format:
   VERDICT: <label>
   REASONING: <one or two sentences citing key evidence>
\end{Verbatim}
\end{tcolorbox}

\begin{tcolorbox}[breakable, colback=gray!5,colframe=blue!70,title=ReAct User Prompt, fonttitle=\bfseries]
\footnotesize
\begin{Verbatim}[breaklines,breakanywhere,showtabs=false,breaksymbol=]
Verify this scientific claim:

{claim}
\end{Verbatim}
\end{tcolorbox}

\begin{tcolorbox}[breakable, colback=gray!5,colframe=blue!70,title=FIRE System Prompt, fonttitle=\bfseries]
\footnotesize
\begin{Verbatim}[breaklines,breakanywhere,showtabs=false,breaksymbol=]
You are a fact-checking agent responsible for verifying the accuracy of claims.
\end{Verbatim}
\end{tcolorbox}

\begin{tcolorbox}[breakable, colback=gray!5,colframe=blue!70,title=FIRE Final Answer Or Next Search Prompt, fonttitle=\bfseries]
\footnotesize
\begin{Verbatim}[breaklines,breakanywhere,showtabs=false,breaksymbol=]
Instructions:
1. You are provided with a STATEMENT and relevant KNOWLEDGE points.
2. Based on the KNOWLEDGE, assess the factual accuracy of the STATEMENT.
3. Before presenting your conclusion, think through the process step-by-step. 
   Include a summary of the key points from the KNOWLEDGE as part of your reasoning.
4. If the KNOWLEDGE allows you to confidently make a decision, output the final 
   answer as a JSON object in the following format:
   {
      "final_answer": "SUPPORT" or "REFUTE" or "UNCERTAIN"
   }
    - "SUPPORT" -- The retrieved evidence contains statements that directly corroborate the claim. The evidence, taken at face value, is sufficient to conclude that the claim is true or highly likely true.
    - "REFUTE" -- Either (a) the retrieved evidence contains statements that directly contradict the claim, or (b) given the scope of the retrieved corpus, a thorough search yields no evidence that substantiates the claim. In both cases, the evidence base does not support accepting the claim as true.
    - "UNCERTAIN" -- The retrieved evidence is relevant to the claim but is ambiguous, incomplete, or internally conflicting such that neither a clear supportive nor a clear refutatory conclusion can be drawn. This includes cases where evidence partially supports the claim but with meaningful caveats, or where sources of comparable credibility disagree.

5. If the KNOWLEDGE is insufficient to make a judgment, issue ONE PubMed and Semantic Scholar search 
   query that could provide additional evidence. Output the search query in JSON 
   format, as follows:
   {
     "search_query": "Your PubMed and Semantic Scholar search query here"
   }
6. The query should aim to obtain new information not already present in the 
   KNOWLEDGE, specifically helpful for verifying the STATEMENT's accuracy.

KNOWLEDGE:
[KNOWLEDGE]

STATEMENT:
[STATEMENT]
\end{Verbatim}
\end{tcolorbox}

\begin{tcolorbox}[breakable, colback=gray!5,colframe=blue!70,title=OpenScholar Claim Framing, fonttitle=\bfseries]
\footnotesize
\begin{Verbatim}[breaklines,breakanywhere,showtabs=false,breaksymbol=]
Is the following scientific claim supported or refuted by the literature: {claim}? Answer uncertain otherwise.
\end{Verbatim}
\end{tcolorbox}

\begin{tcolorbox}[breakable, colback=gray!5,colframe=blue!70,title=OpenScholar Claim Verdict Question, fonttitle=\bfseries]
\footnotesize
\begin{Verbatim}[breaklines,breakanywhere,showtabs=false,breaksymbol=]
Given a scientific question about whether a claim is supported or refuted and a set of retrieved reference passages, output a verdict as valid JSON with exactly these keys:
{
  "label": "SUPPORT" | "REFUTE" | "UNCERTAIN",
  "reasoning": "<one or two sentences citing specific reference numbers>",
  "evidence": ["[0]", "[1]"]
}
Verdict definitions:
- SUPPORT: The retrieved passages contain statements that directly corroborate the claim. The evidence, taken at face value, is sufficient to conclude that the claim is true or highly likely true.
- REFUTE: Either (a) the retrieved passages contain statements that directly contradict the claim, or (b) given the scope of the retrieved corpus, a thorough reading yields no evidence that substantiates the claim. In both cases the evidence base does not support accepting the claim as true.
- UNCERTAIN: The retrieved passages are relevant to the claim but are ambiguous, incomplete, or internally conflicting such that neither a clear supportive nor a clear refutatory conclusion can be drawn. This includes cases where evidence partially supports the claim but with meaningful caveats, or where sources of comparable credibility disagree.
Rules:
- Base your verdict ONLY on the provided references. Do not use outside knowledge.
- If a passage explicitly states the opposite of the claim, that is strong evidence for REFUTE.
- If no passage mentions the entities or relationship in the claim, prefer REFUTE over UNCERTAIN.
- Output valid JSON only. No markdown fences, no extra keys, no prose outside the JSON object.
Question: Is the following scientific claim supported or refuted by the literature: {claim}? Answer uncertain otherwise.
\end{Verbatim}
\end{tcolorbox}

\end{document}